%% file: main_aaai25.tex
\documentclass[letterpaper]{article} 
\usepackage{aaai25}  
\usepackage{times}  
\usepackage{helvet}  
\usepackage{courier}  
\usepackage[hyphens]{url}  
\usepackage{graphicx} 
\usepackage{natbib}  
\usepackage{caption} 
\usepackage{algorithm}
\usepackage{algorithmic}

\usepackage{multirow}
\usepackage{subfigure}
\usepackage{subcaption}

\usepackage{newfloat}
\usepackage{listings}
\DeclareCaptionStyle{ruled}{labelfont=normalfont,labelsep=colon,strut=off} 
\floatstyle{ruled}
\newfloat{listing}{tb}{lst}{}
\floatname{listing}{Listing}
\newcommand{\ny}{\tilde{y}}
\newcommand{\nY}{\widetilde{Y}}
\input{math_commands}

\title{Human and AI Perceptual Differences in Image Classification Errors}

\author{
    Minghao Liu\textsuperscript{\rm 1},
    Jiaheng Wei\textsuperscript{\rm 2},
    Yang Liu\textsuperscript{\rm 1},
    James Davis\textsuperscript{\rm 1}
}

\affiliations{
    \textsuperscript{\rm 1} Department of Computer Science and Engineering,
 University of California, Santa Cruz \\
    \textsuperscript{\rm 2} Data Science and Analytics Thrust, Hong Kong University of Science and Technology (Guangzhou) \\
    mliu40@ucsc.edu, jiahengwei@hkust-gz.edu.cn, yangliu@ucsc.edu, davisje@ucsc.edu
}

\begin{document}

\maketitle

\begin{abstract}

 Artificial intelligence (AI) models for computer vision trained with supervised machine learning are assumed to solve classification tasks by imitating human behavior learned from training labels. Most efforts in recent vision research focus on measuring the model task performance using standardized benchmarks such as accuracy. However, limited work has sought to understand the perceptual difference between humans and machines. To fill this gap, this study first analyzes the statistical distributions of mistakes from the two sources and then explores how task difficulty level affects these distributions. We find that even when AI learns an excellent model from the training data, one that outperforms humans in overall accuracy, these AI models have significant and consistent differences from human perception. We demonstrate the importance of studying these differences with a simple human-AI teaming algorithm that outperforms humans alone, AI alone, or AI-AI teaming.    

\end{abstract}

\section{Introduction}

\noindent
One motivation of neural networks (NN) is creating artificial intelligence that can learn from human intelligence and mimic human behavior. In computer vision, researchers often build their work upon the assumption that neural networks learn a feature representation similar to visual cortex activity \cite{agrawal2014pixels, mur2013human,kuzovkin2018activations}. It is believed that a well-trained machine network learns to represent input stimuli in a way that is similar to human visual perception \cite{battleday2020capturing}. As a result, most current work in computer vision that aims to develop ``better'' computer models focuses on summary benchmark scores (e.g., prediction accuracy).  As a result, the evaluation of human-to-machine similarity across the entire distribution of answers is ignored. 

Human-centric studies do exist in some areas of AI research. Human-centric explainable AI conducts user studies to identify the ``explanations'' that are most meaningful to human users. Noisy label researchers utilize human disagreements as soft labels to train neural networks \cite{branson2010visual,kovashka2016crowdsourcing,peterson2019human,wei2023aggregate}. Human-in-the-loop work utilizes active learning to iteratively expand or update the training set by using human annotators \cite{han2021iterative,joshi2009multi,yao2012interactive,taleb2021multimodal,roels2019human}, or utilizing human feedback reinforcement learning methods to bridge the performance gap between humans and machines \cite{yang2024tackling,ouyang2022training,2016arXiv160303151S,ramakrishnan2019overcoming}. Human workers have also been requested to diagnose AI system failures given semantic explanations of the AI decision \cite{Yang2024, nushi2018towards,nushi2017human}. However, these works focus on developing training schemes to improve the model performance, and limited work explores the perceptual difference between human and machine classifiers. \footnote{More related works are included in appendix (arXiv version). }


Our work quantifies the perceptual difference between humans and machines in the image classification task. We first analyzed the statistical distribution of mistakes made by human annotators and AI annotators (robust machine learning classifiers), where we found that \emph{machines tend to make similar mistakes to other machines, while mistakes differ from those made by humans}. Next, we attempt to understand the differences using implicit information from machine and human classifiers, such as machine confidence scores, human annotation time, and agreement levels, to rank the difficulty of each classification task. Comparing the performance of humans and machines on various task difficulty levels,  we observe that \emph{despite being worse overall, humans outperform the machines in some conditions, such as when machines have low confidence or low agreement}. Finally, we demonstrate the potential of studying human-to-machine differences by using our findings to show that appropriate human-machine collaboration significantly outperforms collaboration between any two machine classifiers. 

The contributions of this work are: 
\begin{itemize}
    \item A comparison of human-machine perceptual differences on an image classification task.
    \item Showing that understanding perceptual differences can improve overall system accuracy.
\end{itemize}

\section{Preliminary}

\subsection{The Multi-Class Classification Task}

In this work, we focus on the multi-class image classification task. Assume that labels  $y \in [1,2,..., K]$ are generated by a random variable $Y$. Ideally, a machine learner would have access to $N$ training data points $D:=\{(x_i,y_i)\}_{i \in [N]}$, where each image/instance $x_i$ generates according to the random variable $X$. The goal of this classification task is to find the optimal classifier $f^{}$ by solving the optimization task:
\[f^*\leftarrow \argmin_{f} \frac{1}{N}\sum_{i\in[N]} \ell(f(x_i), y_i),\]
where $f$ is the classifier (e.g., the machine) and $\ell$ is the evaluation measure (i.e., loss function).

However, labels are often obtained through a crowdsourcing platform (e.g., Amazon Mechanical Turk \cite{mturk}) in practice, which can inevitably incur label/annotation errors to certain images \cite{xiao2015learning,peterson2019human,wei2022learning}. In other words, the learner can only access a set of noisy labels $\ny_i$ generated by the random variable $\nY$, which may disagree with the clean label $y_i$.

\begin{figure}[t]
    \begin{center}
    \includegraphics[width=0.46\textwidth]{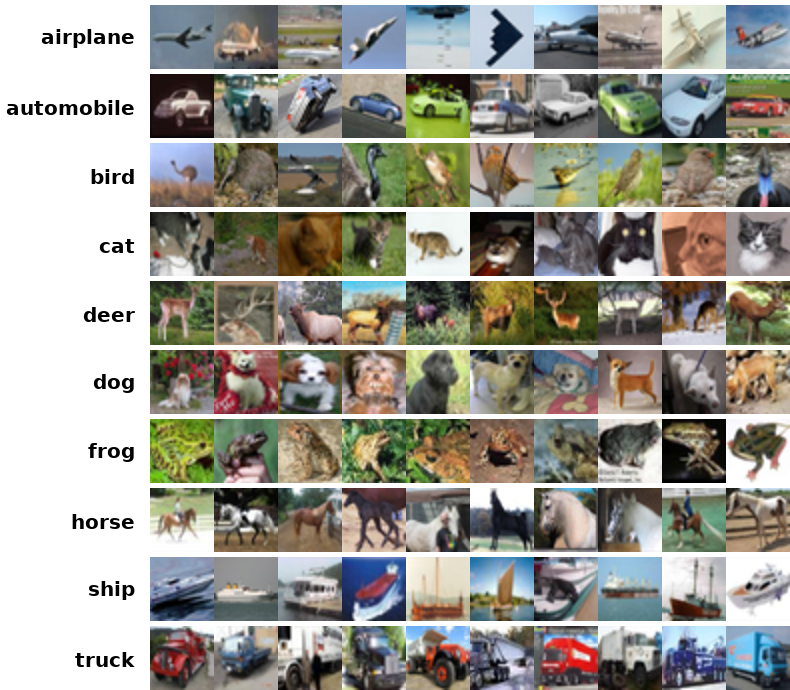}\textbf{}
    \end{center}
    \caption{\textbf{Stimuli for current study:} Representative images from the CIFAR-10 dataset, which includes ten categories of natural images. The perceptual difference between human and AI classifiers is studied using the distribution of mistakes made while predicting categorical labels.   }
    \label{Evidence cifar-fig}
\end{figure}

\subsection{Datasets and Model Training}
\noindent
The top-ranked machine vision models can achieve extremely high accuracy on CIFAR-10 \cite{krizhevsky2009learning} image classification by training on clean labels. The models are so good with clean labels that this dataset is often considered a toy problem domain. However, label noise is prevalent in real-world classification data \cite{deng2009imagenet,xiao2015learning,wei2022learning}. Training models on datasets with label noise inevitably results in biased model prediction \cite{natarajan2013learning,liu2015classification,patrini2017making}. Thus CIFAR-10 remains a widely used test dataset when studying image classification using noisy training data \cite{liu2020peer,cheng2020learning,liu2020early,wei2021when,liu2022robust,wei2022open,wang2022promix,huang2023paddles,xia2023regularly,ortiz2023does,chen2023understanding,chen2023imprecise,park2024robust,wang2024binary}. The machine classifier must decide which of ten class labels (airplane, automobile, bird, cat, deer, dog, frog, horse, ship, truck) best describes each image.   
An example of images in this dataset is shown in Figure~\ref{Evidence cifar-fig}.

In this paper, we adopt CIFAR-N \cite{wei2022learning}, a label-noise benchmark that provides three noisy human annotations for each image of the CIFAR-10 training dataset. We split the training set into a 40K training subset and a 10K test subset. We explore human perceptual differences using these noisy human annotations. For machine classifiers, there exists a family of robust designs for learning with noisy labels. We choose thirteen popular robust methods, allowing the machine to mitigate the impact of noisy labels.

A preliminary comparison of human and machine classifier accuracy is given in Table~\ref{tab:acc}. Noisy human annotators from CIFAR-N achieve 81.9-82.8\% accuracy. However, all the selected machine classifiers achieved higher accuracies than humans, ranging from 84.5\% to 89.9\%, when trained using data from CIFAR-N. 
\begin{table}[t]
\centering
\scalebox{0.9}{
    \begin{tabular}{cc|c}
    
    \hline
    \multicolumn{2}{l|}{}                                         & Accuracy (\%)\\ \hline
    \multicolumn{1}{c|}{\multirow{3}{*}{Humans}}     & Human 1     & 82.8     \\ 
    \multicolumn{1}{c|}{}                           & Human 2      & 81.9     \\ 
    \multicolumn{1}{c|}{}                           & Human 3     & 82.4     \\ \hline
    \multicolumn{1}{c|}{\multirow{13}{*}{Machines}} & CORES \cite{cheng2021learning}       & 84.5     \\ 
    \multicolumn{1}{c|}{}                           & CE          & 85.0     \\ 
    \multicolumn{1}{c|}{}                           & PLS \cite{lukasik2020does}       & 85.9     \\ 
    \multicolumn{1}{c|}{}                           & F-Div \cite{wei2021when}      & 86.0     \\ 
    \multicolumn{1}{c|}{}                           & GCE \cite{zhang2018generalized}         & 86.3     \\ 
    \multicolumn{1}{c|}{}                           & FW \cite{patrini2017making}          & 86.7     \\ 
    \multicolumn{1}{c|}{}                           & PeerLoss \cite{liu2020peer}   & 86.8     \\ 
    \multicolumn{1}{c|}{}                           & BW \cite{patrini2017making}          & 86.8     \\ 
    \multicolumn{1}{c|}{}                           & NLS \cite{wei2022smooth}         & 88.4     \\ 
    \multicolumn{1}{c|}{}                           & CAL \cite{zhu2021second}        & 88.7     \\ 
    \multicolumn{1}{c|}{}                           & Co-teaching+ \cite{yu2019does} & 89.1     \\ 
    \multicolumn{1}{c|}{}                           & JoCoR \cite{wei2020combating}      & 89.5     \\ 
    \multicolumn{1}{c|}{}                           & Co-teaching \cite{han2018co}  & 90.0     \\ \hline
    \end{tabular}
}

\caption{\textbf{Overall image classification prediction accuracies}: Robust machine learning models are trained on noisy human labels in CIFAR-N (40K training subset). The accuracies of humans and machines are calculated against the clean label in CIFAR-10 (10K test subset). Note that all machine models outperform all noisy human annotators.}
\label{tab:acc}
\end{table}
A higher accuracy score is widely believed to imply that machines can replace human annotators in this classification task. Our study investigates: (1) Whether perceptual differences exist between humans and machines; and (2) Are these differences important to the overall machine performance?


\section{Quantifying Perceptual Differences}
\label{sec:diff}

\noindent
In this section, we study the perceptual difference between humans and machines in image classification tasks. Our objective is to determine whether human capabilities are merely a subset of machine abilities or if they can complement each other. To represent their respective perceptual distributions, we use a set of machine predictions and noisy human annotations to conduct our analysis. To quantify and compare the distribution of predictions, Sec \ref{sec:diff-human vs. ML} makes use of confusion matrices to illustrate the pattern of their incorrect predictions, Sec \ref{sec:diff-ml diffculty}-\ref{sec:diff-human diffculty} demonstrates their performance on tasks with various difficulty levels, and Sec \ref{sec:more training scheme} shows additional results when permuting training data and learning backbone.

\subsection{The Prediction Confusion Matrix}
\label{sec:diff-human vs. ML}
To understand the overall perceptual distribution difference between human and machine classifiers, we visualize their incorrect predictions through confusion matrices.

Denote by $f_{\text{ML}}$ a machine learning classifier, and $f_{\text{H}}$ as a human annotator. For $f\in\{f_{\text{ML}}, f_{\text{H}}\}$, let $\mathbf{f}(x)$ be the model prediction probability vector given by the classifier $f$ w.r.t. the sample $x\in X$, i.e., $\mathbf{f}(x)=[\mathbf{f}(x)_{[1]}, ..., \mathbf{f}(x)_{[K]}]$, where $\mathbf{f}(x)_{[i]}:=\mathbb{P}(f(x)=i)$ denotes the probability of categorizing $x$ as class $i$, in a $K$-class classification task. The model prediction could then be expressed as $f(x):=\argmax_{i\in[K]} \mathbf{f}(x)_{[i]}$. Given a set of test data samples with clean labels $D^{\text{t}}:=\{(x_i^{\text{t}},y_i^{\text{t}})\}_{i \in [N]}$, for $f\in\{f_{\text{ML}}, f_{\text{H}}\}$, we aim to quantify the differences between $f_{\text{ML}}$ and $f_{\text{H}}$ through the following metric:

\vspace{0.5em}
\noindent
\textbf{Confusion matrices of model predictions $\mathbf{C}$:} The elements of confusion matrix $\mathbf{C}$ are given by 
    \[\mathbf{C}_{p, q}:=\mathbb{P}_{(X, Y)\sim D^{\text{t}}}(f(X)=q|Y=p).\]
    To quantify the perceptual difference between humans and machines, one could replace the ground-truth (clean) label $Y$ with the human annotation $f_{\text{H}}(X)$ and adopt  $f=f_{\text{ML}}$.

\begin{figure*}[!t]
  \makebox[\textwidth][c]{
      \includegraphics[width=1\textwidth]{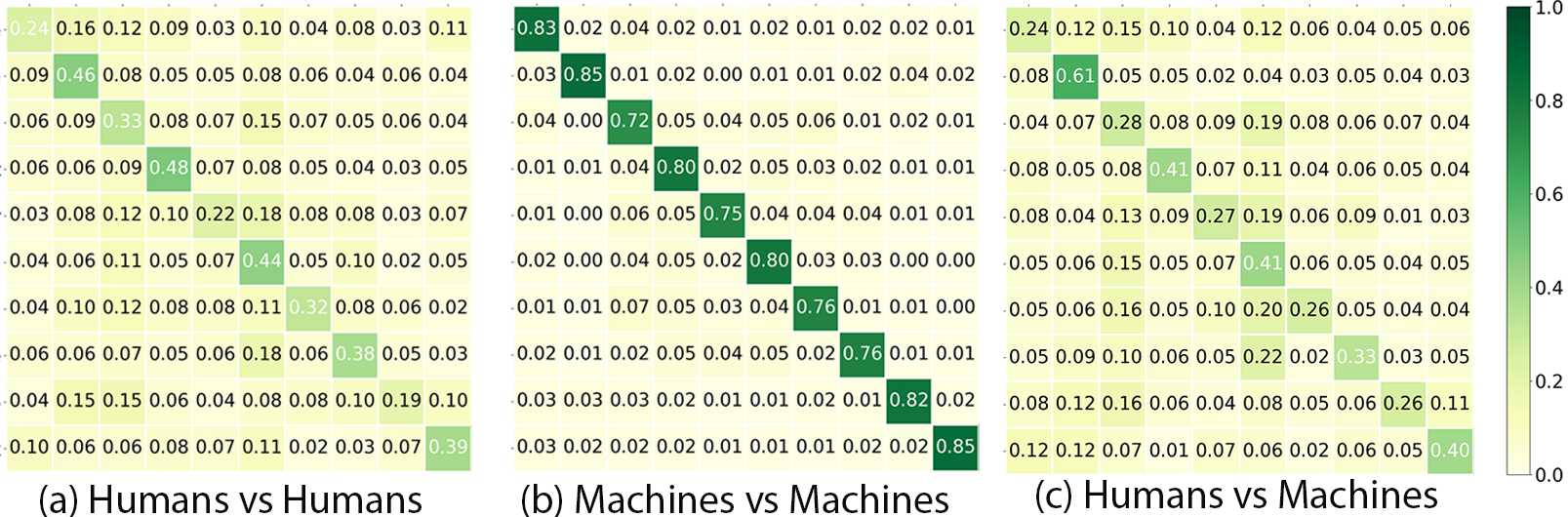}
    }
  \caption{\textbf{Confusion matrices of incorrect answers:} The figure shows confusion matrices across permutations of machine classifiers or human annotators. The plot focuses on incorrect predictions from the test subjects to see if they make similar mistakes. For example, a darker cell in "Humans vs Machines" means a higher probability that the three human annotators make the same mistake as the thirteen designs of machine classifiers. (a) A mild diagonal line indicates that humans don't always make the same mistakes. (b) The strong diagonal line indicates that all the machine models tend to make similar mistakes. (c) The diagonal line is weak, indicating that the mistakes made by humans and machines diverge in this comparison.}
    \label{Evidence 101}
\end{figure*}

 Figure~\ref{Evidence 101} shows the aggregated confusion matrices for samples with incorrect prediction between human-human, human-machine, and machine-machine. For instance, the "Humans v.s. Machines" figure compares three sets of human annotations against all thirteen designs of machine classifiers. By examining these confusion matrices, we can gain insight into the distribution of human and machine mistakes.

\vspace{0.5em}
\noindent
\textbf{Comparisons between machines and machines:} 
Denote by $\mathbf{C}^{\text{MM}}$ the confusion matrix of model predictions between a machine and a machine. Figure~\ref{Evidence 101} (b) shows the aggregated confusion matrix $\mathbf{C}^{\text{MM}}$ between all pairs of machines. We observed a strong diagonal line in the plot, indicating that machines have a strong consensus and tend to make similar mistakes.

\vspace{0.5em}
\noindent
\textbf{Comparisons between humans and humans:} 
We define $\mathbf{C}^{\text{HH}}$ as the confusion matrix of model predictions between a human annotator and another human annotator. The human-human comparison $\mathbf{C}^{\text{HH}}$ is shown in Figure~\ref{Evidence 101} (a). The diagonal line is weaker since human judgments sometimes diverge. By comparing $\mathbf{C}^{\text{HH}}$ and $\mathbf{C}^{\text{MM}}$ (Figure \ref{Evidence 101} a,b), we conclude that although humans in our test set have similar accuracies, they sometimes make different mistakes. In contrast, machines models, even those with different accuracies, tend to make similar mistakes.

\vspace{0.5em}
\noindent
\textbf{Comparisons between machines and humans:} Denote by $\mathbf{C}^{\text{HM}}$ the confusion matrix of model predictions between a human annotator and a machine. 
 The human-machine comparison $\mathbf{C}^{\text{HM}}$ is shown in Figure~\ref{Evidence 101} (c). By comparing the confusion matrices $\mathbf{C}^{\text{MM}}$ vs. $\mathbf{C}^{\text{HM}}$ (Figures~\ref{Evidence 101} b,c), we see that humans and machines make significantly different mistakes. 

Hypothesis testing is utilized to support our visual observations. With a significance level of $\alpha=0.05$, the hypothesis testing results showed there exist significant differences between $\mathbf{C}^{\text{HH}}$ vs. $\mathbf{C}^{\text{MM}}$ and $\mathbf{C}^{\text{MM}}$ vs. $\mathbf{C}^{\text{HM}}$. Details are provided in Appendix B.1. 


The perceptual difference between humans and machines suggests human capabilities aren't strictly a subset of machine abilities, implying the potential for complementary teaming.

\subsection{Partitioning by Machine Difficulty}
\label{sec:diff-ml diffculty}

\noindent
To better understand perceptual differences, we partition test images into subsets that may have performance variations. As a first attempt at partitioning, we group test cases based on the difficulty level for machine learning models $f_{\text{ML}}$ to provide an accurate answer. The performance of humans and machines is evaluated for each subset. We try two difficulty measures: the machine's model confidence and agreement.

\vspace{0.5em}
\noindent
\textbf{{Machine's model confidence:}}
Define model confidence as $\text{MC}(f=f_{\text{ML}}|x):=\mathbf{f_{\text{ML}}}(x)_{f_{\text{ML}}(x)}$ or $\argmax_{j\in[K]} \mathbf{f_{\text{ML}}}(x)_{j}$, which indicates the model's prediction probability on sample $x$ for its prediction $f_{\text{ML}}(x)$.

\begin{figure*}[!t]
    \centering
     \includegraphics[width=1.0\textwidth]{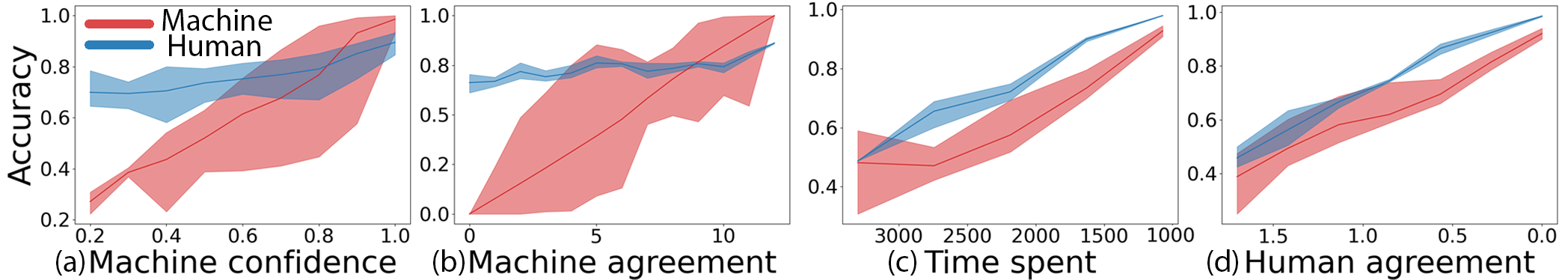}\textbf{}
    \caption{\textbf{Accuracy as a function of difficulty level:} The plots visualize the performance of humans and machines on tasks ranked by difficulty level. The shaded band indicates the range of accuracies for all classifiers, and the solid line represents the average performance.  Task difficulty is measured by: \textbf{(a)} machine classifier confidence levels, \textbf{(b)} based on machine agreements, \textbf{(c)} based on human annotation time, \textbf{(d)} and human agreement levels. Plots (a,b) show machine classifier performance heavily correlates to machine difficulty levels, while human performance is significantly less correlated. Plots (c,d) indicate that both human and machine performance is correlated to human-derived difficulty levels.}
    \label{Evidence 102}
\end{figure*}

\vspace{0.5em}\noindent
The machine confidence score is a commonly used metric to show the model prediction certainty level. A higher confidence score indicates a higher level of certainty and suggests that the task is easier for the machine to perform. In our experiments, we calculated the accuracy of humans and machines at different confidence levels using each machine classifier. Results are plotted in Figure~\ref{Evidence 102} (a). The shaded band indicates the range of accuracies, and the solid line represents the average.  The performance of the thirteen machine designs is shown in red, while the results of human classifiers are shown in blue.

Examining the plot, only the red band shows a strong upward trend when the confidence level increases, meaning that the machine performance strongly correlates to the machine confidence score. In contrast, human performance is only mildly correlated with machine confidence. In addition, the plot shows that humans are more accurate than machines when machines have low confidence. Even though machines have higher accuracy on the complete test set, humans and machines have different expertise on different subsets, neither can fully replace the other on this classification task.

\vspace{0.5em}
\noindent
\textbf{Machine agreement:}
Given a list of machine learning classifiers, i.e., $k$ classifiers with $j$-th one denoted by $f_{\text{ML}, j}$, we could quantify the difficulty level for machines w.r.t. a sample $x$ as $\text{MA}(f=f_{\text{ML}}|x):=\frac{1}{k}\sum_{j\in [k]} \1(f_{\text{ML}, j}(x)=y)$. The model agreement term $\text{MA}(f=f_{\text{ML}}|x)\in [0,1]$ indicates a high consensus rate for machine learning classifiers (low difficulty level) if it is of a high value.

\noindent
Machine agreement uses a cluster of machine classifiers to rank the task difficulty. More machines making correct judgments on a task indicates the task is easier to solve, and vice versa. We repeat the experiment with a leave-one-out design, using 12 machine classifiers to determine the difficulty and one remaining classifier to test. We experimentally compare human and machine accuracies on tasks with different machine agreement levels, then visualize the difference in Figure~\ref{Evidence 102}~(b). The result again shows that machine performance strongly correlates with other machine classifiers, even though various machine designs have a wide range of performance. Human annotator performance, on the other hand, is only slightly correlated to machine agreement. Here, we draw a similar conclusion to using machine confidence levels: human and machine classifiers have different distributions of predictions.

\vspace{0.5em}
\noindent
\textbf{Statistical hypothesis testing:} We verify our visual observations using a linear regression fit, which shows the correlation between the accuracy and task difficulty levels. A larger slope indicates a stronger relation, while values closer to zero indicate a low correlation. We set the significance value to be $\alpha=0.05$: the average of human classifiers has a slope of 0.11 w.r.t. the machine confidence and a slope of 0.08 with machine agreements, while machines have a slope of 1.00 in both cases. This illustrates that the difficulty level, as judged by machines, has a low correlation with human predictions. Details are provided in Appendix B.2, B.3.


\subsection{Partitioning by Human Difficulty}
\label{sec:diff-human diffculty}

\noindent
We repeat the partitioning experiments via new criteria: this time partitioning by human difficulty rather than machine difficulty. To quantify the human difficulty levels, we use time spent labeling an image and an entropy-based measure of human agreement.

\vspace{0.5em}
\noindent
\textbf{Time Spent:}
Assuming the $i$-th human annotator spent $t_i(x)$ time (in seconds) in annotating sample $x$, we adopt the average time spent on sample $x$ as a measure to indicate the difficulty level of the given task $x$. Mathematically, $\bar{t}(x):=\frac{1}{k}\sum_{i\in[k]} t_i(x)$. A large $\bar{t}(x)$ means the sample $x$ is relatively hard for human annotators since it requires humans to spend a long time on annotation. We calculated average time consumption using the CIFAR-H dataset~\cite{peterson2019human} since CIFAR-N lacks annotator time information.

\vspace{0.5em}
\noindent
\textbf{Human agreement (entropy):}
Given $k$ human annotators, the entropy of a sample $x$ is given by: \begin{align*}
    \text{Entropy}(x)=-\sum_{i\in[K]}p_{\text{H}, i}(x)\cdot \log{\left(p_{\text{H}, i}(x)\right)},
    \end{align*} 
where $p_{\text{H}, i}(x):=\frac{1}{k}\sum_{j\in [k]} \1(f_{\text{ML}, j}(x)=i).$

\noindent
Human agreement is a metric to evaluate human consensus on a task given multiple annotations. A higher agreement level implies less ambiguity and easier judgment. We calculate entropy using CIFAR-H because it provides at least 47 labels per image, enough to calculate agreement.

\vspace{0.5em}
\noindent
\textbf{Analysis:}
We calculated human and machine performance under different human difficulty levels, then aggregated results as shown in Figure~\ref{Evidence 102} (c,d). Both human and machine performance strongly correlate with annotation time and human agreement. Here, we conclude that machines will struggle with cases humans find difficult. While this result contrasts our findings using machine difficulty to partition tasks, it does not contradict those results. Machine and human difficulty metrics result in different partitions, and the statistics of these partitions are different.

\subsection{Sensitivity of the Results}
\label{sec:more training scheme}

\noindent
Comparing human and machine classifiers, we have found that machines are more similar to other machines and that humans have different statistics. This result may come from using the same training data for all machine classifiers or, alternately, from using the same vision backbone on all classifiers. To understand whether this is the cause, we conduct further experiments to provide variation to the machine classifiers during the training phase.

 To ensure fairness in this test, we created a balanced test set to reduce the bias potentially caused by the majority of correct cases. A balanced set is a subset randomly selected from the test set, which includes half correct and half incorrect answers. To measure the similarity of two classifiers, we used \textit{Matching percentage on a balanced set} as the numerical metric, defined as $\frac{P(A|B) +P(\overline{A}|\overline{B})}{2}$ for two sets $A, B$.

\begin{figure*}[btp]
    \centering  
    \includegraphics[width=1.0\textwidth]{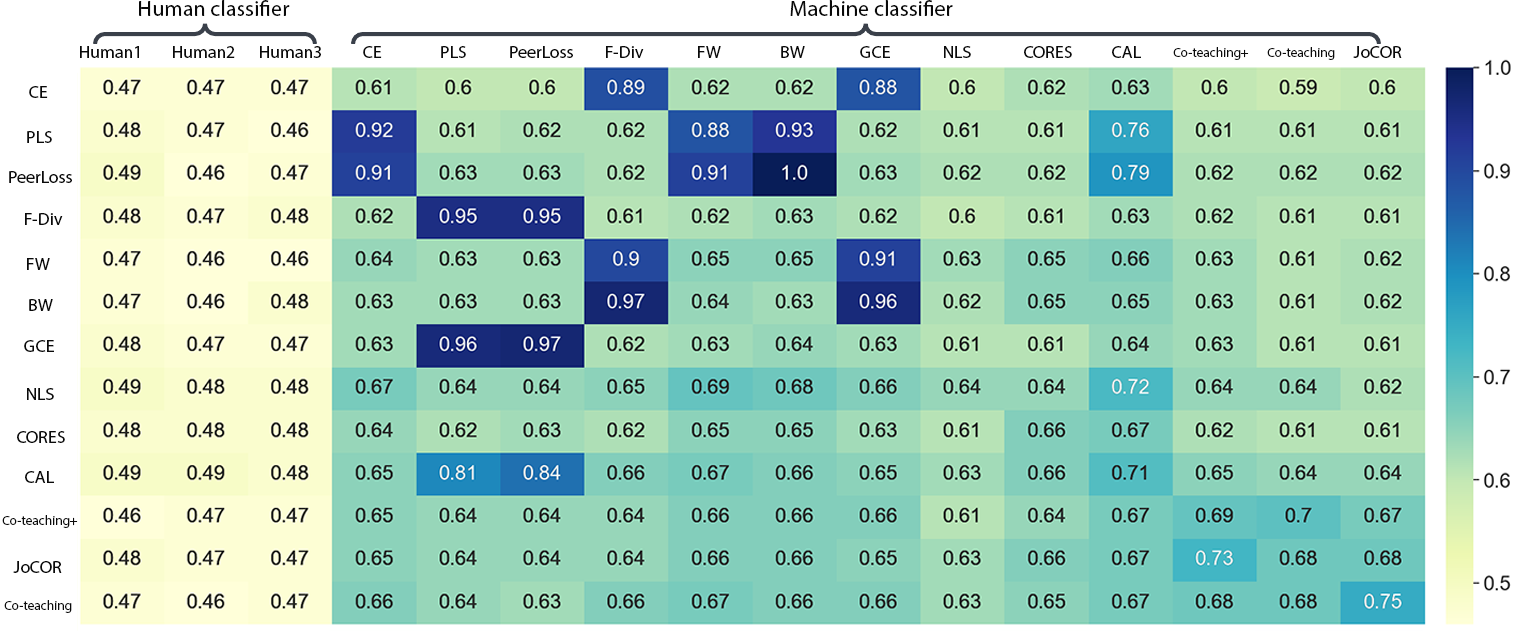}\textbf{}

    \caption{\textbf{Matching Percentage on balanced set:} The figure visualizes the matching percentage between each machine classifier and each other human/machine classifier on a balanced set. The machine classifiers are not all trained with the same training examples, yet the results show machines tend to make judgments that match other machines more than they match humans. }

    \label{Evidence 106}
\end{figure*}

\vspace{0.5em}
\noindent
\textbf{Training labels:} We diversify the training labels to evaluate whether using the same training labels is the cause of different machine classifiers making similar mistakes. We train each machine on one of the three human annotations sets from the CIFAR-N dataset and compare their Matching percentage on a balanced test set in Figure~\ref{Evidence 106}. Higher matching percentages mean the two classifiers are more similar (darker colors). Notice that machines and humans tend to make less similar judgments, while machines tend to make similar judgments to other machines, even when trained with different sets of labels.

\begin{table*}[h]
\centering{\scalebox{0.86}{
\begin{tabular}{c|cc|cc|cc}
 \hline
Pre-trained& \multicolumn{2}{c}{ $\text{Acc}_{\text{col, ml}}(0.4) \text{ v.s. }\text{Acc}^*_{\text{col, ml}}$}& \multicolumn{2}{c}{ $\text{Acc}_{\text{col, ml}}(0.5) \text{ v.s. }\text{Acc}^*_{\text{col, ml}}$ } & \multicolumn{2}{c}{ $\text{Acc}_{\text{col, ml}}(0.6) \text{ v.s. }\text{Acc}^*_{\text{col, ml}}$}  \\
\hline
& Statistic&  $p$-value & Statistic&  $p$-value & Statistic&  $p$-value\\ 
&-18.96  & 7.51e-6  &  -7.95 & 0.0005 & -2.49  & 0.0555 \\
 \hline\hline
Multi-nets& \multicolumn{2}{c}{ $\text{Acc}_{\text{col, ml}}(0.4) \text{ v.s. }\text{Acc}^*_{\text{col, ml}}$}& \multicolumn{2}{c}{ $\text{Acc}_{\text{col, ml}}(0.5) \text{ v.s. }\text{Acc}^*_{\text{col, ml}}$ } & \multicolumn{2}{c}{ $\text{Acc}_{\text{col, ml}}(0.6) \text{ v.s. }\text{Acc}^*_{\text{col, ml}}$}  \\
\hline
& Statistic&  $p$-value & Statistic&  $p$-value & Statistic&  $p$-value\\ 
& -66.25 & 0.0002  &  -37.57 & 0.0007 & -34.12  & 0.0009 \\
 \hline
\end{tabular}}}
\caption{Results of hypothesis test2 with paired student t-test: we adopt the significance value $\alpha=0.05$.}
\label{tab:test_collabrate_main}
\vspace{-0.1in}
\end{table*}

\vspace{0.5em}
\noindent
\textbf{Vision backbones:} We selected three commonly used vision backbones: ResNet~\cite{he2016deep}, VGG~\cite{simonyan2014very}, and Inception~\cite{szegedy2015going}. The machine classifiers were retrained, each using a randomly selected backbone. The matching percentage was again used as a metric for similarity. For space reasons, the table of numbers from this experiment is in the supplementary materials (Appendix C.1). However, we draw the same conclusion. Machines and humans make different judgments, while two machine classifiers tend to make similar judgments, even when trained with different vision backbones.



\section{Human-Machine Collaboration}
\label{sec:collaborate}

\noindent
We investigate the performance of joint systems using human and machine statistical distributions. Although machines have higher overall accuracies in our study, there exist cases in which humans perform better. The goal of this section is to demonstrate that the study of perceptual differences has practical value. We thus choose the simplest human-machine collaboration method possible, in which a joint classifier has access to both a machine guess and a human guess and returns one of these two answers. We investigate an upper bound on performance using an ideal collaboration and obtainable performance using a realistic collaboration based on simple thresholding.

\subsection{Ideal Collaboration}

\begin{figure}[!t]
    \begin{center}

    \includegraphics[width=0.46\textwidth]{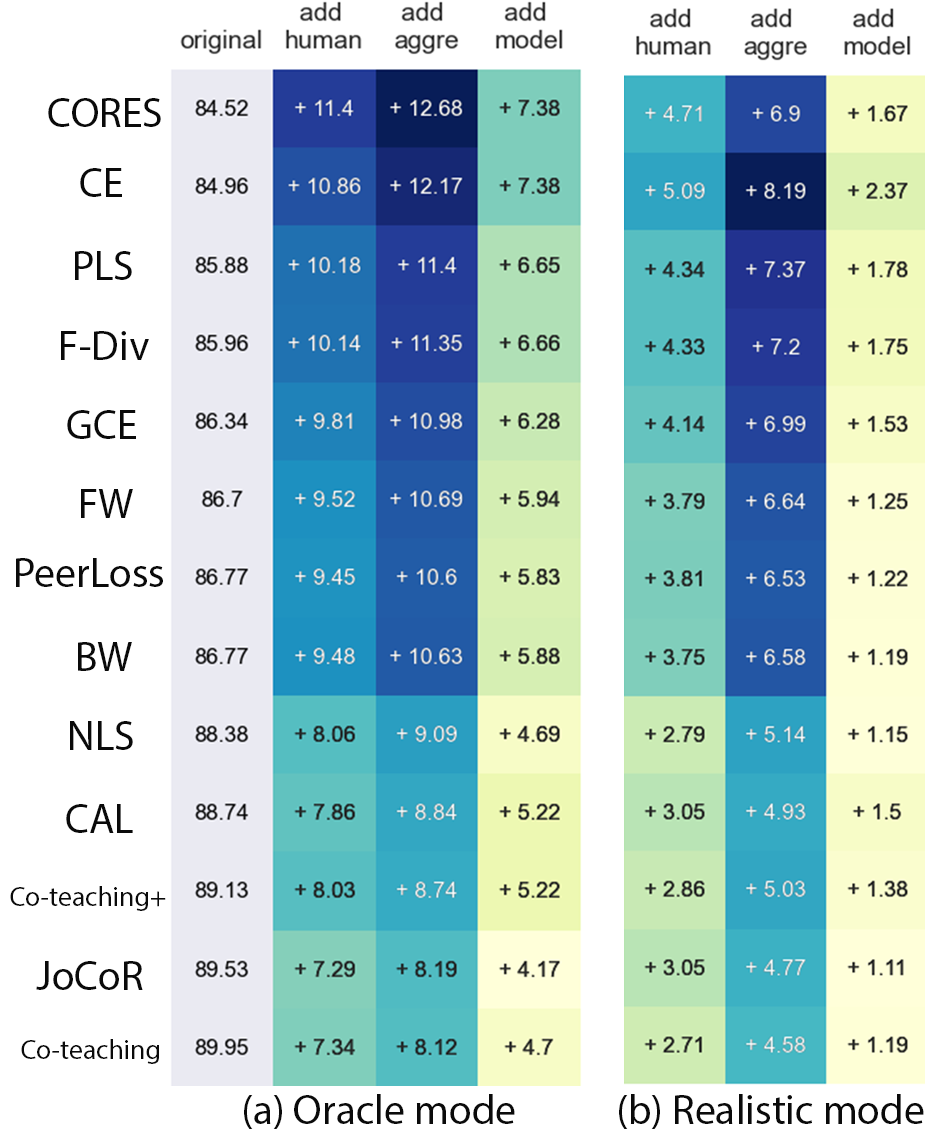}\textbf{}
    \end{center}
    \caption{\textbf{Post-hoc teaming:} The figure shows the original model performance and the boost from teaming options. \emph{``Add human"} is teaming with a human classifier, \emph{``Add aggre"} is teaming with a human classifier that aggregates answers from three humans, and \emph{``Add model"} is teaming with another machine classifier. We compared all the permutations and visualized the best teaming combinations using a color map. A darker color indicates a greater boost. \textbf{(a)} Oracle mode is the upper bound from perfect teaming, \textbf{(b)} realistic mode is from a simplistic real algorithm. The results show the value of human-machine complementary teaming. Introducing a low-performance human to the teaming system causes more boost than introducing a higher-performance machine classifier.}
    \label{Evidence 201}
\end{figure}

\noindent
 Given predictions from a machine classifier and a human classifier, ideal collaboration selects the correct answer if either classifier generates it. Algorithms that use ground truth data to make perfect predictions are frequently called ``oracles" in the computer vision literature, and we adopt this terminology, labeling ideal collaboration ``oracle mode." If either classifier predicts the correct result, oracle mode will have the correct result. This is an upper bound for what is achievable in a real system.

Oracle mode performance is given in Figure \ref{Evidence 201}(a), which shows the original machine classifier performance and the performance boost obtained from teaming. We show three options for teaming with each original machine predictor: teaming with a human classifier, teaming with an aggregated group of three human classifiers, and teaming with another machine classifier. The additional machine classifier is chosen by examining all options and reporting the pair with the greatest accuracy boost.

In all cases, teaming with a human has a greater upper bound on accuracy gain than teaming with another machine model. Using the machine cross-entropy (CE) classifier as an example, we see that it has an accuracy of {\small{\tt{{(84.96\%)}}}}. When teaming with a human classifier{\small{\tt{{(81.88\%)}}}}, the joint system can reach {\small{\tt{{95.82\%}}}}{\small{\tt{{(+10.86\%)}}}} under this ideal collaboration condition. The boosted accuracy is significantly greater than each classifier alone. On the other hand, teaming the CE classifier with another machine classifier such as coteaching\cite{han2018co} {\small{\tt{{(89.95\%)}}}}  results in a teaming performance of {\small{\tt{{92.34\%}}}}{\small{\tt{{(+7.38\%)}}}}. Note that the lower accuracy human classifier results in better teaming performance than the higher accuracy co-teaching model. Indeed, the machine-machine teaming accuracy is only mildly better than co-teaching accuracy alone. 

To investigate whether multiple human classifiers can further improve performance, we aggregate annotations from three human annotators using the majority vote to select the classification. This \textit{aggre} classifier {\small{\tt{(90.97\%)}}}, has an accuracy much greater than a single human {\small{\tt{{(81.88\%)}}}}, but under the oracle teaming condition boosts accuracy only a little more to {\small{\tt{{97.13\%}}}}{\small{\tt{{(+12.17\%)}}}}.

The upper bound of teaming between machine and human classifiers outperforms any combination of machine-machine classifiers, revealing the benefit of leveraging human-machine complementary characteristics.

\subsection{Realistic Collaboration}

To understand performance in a realistic scenario without access to an oracle, we evaluate a simple algorithm that thresholds machine confidence to determine whether to use the machine or human guess. 

The early memorization effect inspires thresholding. The learning-with-noisy-label literature~\cite{liu2020early,wei2022learning} observes that deep neural nets tend first to memorize clean patterns, then gradually fit on noisy patterns as training progresses. Samples that are of bad patterns usually have large losses, which degrades the generalization of the model. When evaluating the model performance on the test data, we would expect test samples that are similar to the learned training samples with clean labels to have relatively high confidence. Inspired by this effect, we replace low-confidence machine model predictions with those from a human classifier while keeping confident model predictions from the original model. This is not guaranteed to boost accuracy since the human classifier has lower overall accuracy, and sometimes, a correct classification will be replaced with an incorrect one. However, we saw previously that model confidence is strongly correlated with model accuracy, so we have reason to believe that the approach works.

We adopt a threshold value $\eta \in [0,1]$ to determine when to swap the predictions of the machine predictors with the human classifier predictions. 
Selecting $\eta$ to be 0 or 1 would use either all predictions from the machine predictors or all from the human classifier. To select the optimal threshold value, we use the paired student t-test to compare the algorithm's accuracy with different threshold values. The test shows that swapping model predictions with confidence $\eta\leq 0.6$ results in significant accuracy improvements, with a significance value less than $p=0.05$. More details about the t-test and its results can be found in next section.


Fig \ref{Evidence 201}(b) shows the performance boost under the same three teaming options used previously. Teaming with a human classifier produces a greater boost to accuracy than teaming with the best-choice machine model. All combinations of machine-machine teaming lead to only marginal boosts in accuracy. Using additional human classifiers in the \textit{aggre} majority voting configuration leads to additional gains. 

 Since the upper bound performance is well above the simple realistic algorithm used, it seems likely that better real collaboration algorithms are possible.

\vspace{0.5em}
\noindent
\textbf{Picking the threshold for human-machine collaboration}

Given a confidence threshold $\eta$, we denote by $\text{Acc}_{\text{col, ml}}(\eta)$ as the test accuracy of replacing unconfident model predictions trained with a specific method by human annotations. We adopt paired t-test to see: for all robust methods, whether replacing unconfident model predictions (w.r.t. a fixed $\eta^*$) by human annotations could be consistently the best among such a human-machine collaboration (i.e., $\text{Acc}^*_{\text{col, ml}}$). Mathematically, 
\begin{align*}
&\mathbf{H_0}: \text{ There are no significant differences between }\\
&\hspace{1em}\text{Acc}_{\text{col, ml}}(\eta^*) \text{ and }\text{Acc}^*_{\text{col, ml}},\text{ for a list of robust methods};\\
&\mathbf{H_1}: \text{ There are significant differences between }\\
&\hspace{1em}\text{Acc}_{\text{col, ml}}(\eta^*) \text{ and }\text{Acc}^*_{\text{col, ml}},\text{ for a list of robust methods}.
\end{align*}
 
The hypothesis testing results are summarized in Table~\ref{tab:test_collabrate_main}. We are interested in two lines of methods: methods require a pre-trained model or multiple networks. For the pre-trained model, we observe that the $p$-value for $\eta=0.6$ satisfies that $p\geq \alpha=0.05$, indicating that we should accept the null hypothesis $\text{H}_0$, and conclude that there exists no significant differences between $\text{Acc}_{\text{col, ml}}(\eta^*=0.6) \text{ and }\text{Acc}^*_{\text{col, ml}}.$

\section{Limitations}

We chose CIFAR-10 for our study because it's widely recognized and gives us a common ground for comparison. However it's possible that our findings here might not be generalizable to other datasets. In addition, our study necessarily looked at only some of the many machine classification methods. We hope that demonstrating that human machine perceptual differences exist in at least some cases is sufficient to motivate future research on additional data and methods.
\section{Conclusions}

In this work, we show that perceptual differences between human and machine classifiers exist on image classification tasks. Our analysis shows that while humans have lower overall accuracy than machines in our task, their perception is not a mere subset. All machine classifiers in our tests were strongly correlated with each other, while human classifiers produced a different distribution of answers. Given such differences in expertise, we show that human-machine teaming can lead to greater accuracy than machine-machine teaming.

This work has several practical implications. For scientists seeking to use AI as a digital twin for studying human decision-making, this work shows a weakness in current machine classification models. For application engineers with high-value image classification tasks, such as detecting cancer in medical images, this work demonstrates that human-machine collaboration may result in total system accuracy better than what is achievable by machines or humans alone. For computer vision researchers seeking to create fully automated machine classification systems, the findings in this study suggest that tuning algorithms solely for accuracy may not lead to the highest accuracy. If machine classifiers can be created with statistics more like humans, they will initially have lower total accuracy than the current state of the art. However, these new classifiers with differing statistics can then be combined with current methods to produce a fully automated joint system with higher performance than is currently possible.

\section*{Acknowledgements} Partial support for M.L. and J.D. was provided by WISEautomotive through the ATC+ Program award from MOTIE Korea 20014264t
\newpage

\bibliography{ref}

\onecolumn
\appendix
\setcounter{page}{1}
    \section*{\Large Appendix}

\input{appendix/appendix}
\twocolumn 

\end{document}

%% file: math_commands.tex
\usepackage{amsmath,amsfonts,bm}

\def\eqref#1{equation~\ref{#1}}

\def\1{\bm{1}}

\DeclareMathAlphabet{\mathsfit}{\encodingdefault}{\sfdefault}{m}{sl}
\SetMathAlphabet{\mathsfit}{bold}{\encodingdefault}{\sfdefault}{bx}{n}

\DeclareMathOperator*{\argmax}{arg\,max}
\DeclareMathOperator*{\argmin}{arg\,min}

%% file: appendix/appendix.tex

\section{Related work}
\label{sec:related works}
We included a more detailed version of related work than was appropriate in the main text.

\subsection{Algorithmic-centric machine evaluation}

Researchers in the field of computer vision have extensively evaluated and compared algorithms focused on designing high-performing deep-learning models. For example, MLP-Mixer \cite{tolstikhin2021mlp} combines MLP blocks with skip connections, AlexNet stacks multiple convolutional neural networks (CNNs) \cite{krizhevsky2012imagenet,lecun1989backpropagation}, ResNet combines CNNs with forward skip connections \cite{he2016deep}, and transformer models utilizing attention layers\cite{dosovitskiy2020image,vaswani2017attention}.
Numerous studies have compared the performance and learned feature representations of different backbone designs \cite{tolstikhin2021mlp,guo2021beyond}.

Explainable AI has developed tools to understand which parts of an input drive a machine's final decision. Existing methods include gradient-based and attribute-based approaches to explain CNN-based networks, which create heatmaps to show local relevance to the input image. Gradient-based methods utilize the combination of the input activation and gradient with respect to the input of each layer \cite{smilkov2017smoothgrad,selvaraju2017grad}. Attribute-based methods decompose the output of each layer based on the contributions of earlier layers \cite{montavon2017explaining,iwana2019explaining}. Other intuitive methods treat the model as a black box \cite{dabkowski2017real,zhou2018interpreting,erhan2009visualizing}. A comprehensive survey of explainable AI can be found in \cite{alicioglu2022survey}.

Voita \textit{et al.} \cite{voita2019analyzing} applied Layer-wise Relevance Propagation (LRP) to ViT models to visualize the attention heat map. LRP showed the different relevance of the attention head in each layer. Abnar \textit{et al.} \cite{abnar2020quantifying} proposed the roll-out method, which assumes that the attention from each layer is linearly combined with the others. Gradient-based methods have also been used to visualize ViT models \cite{selvaraju2017grad,chefer2021transformer}.

However, this line of work has focused on comparing different machine models or on visualizing the relevance of a single model's input decision using automatic algorithms or metrics. In contrast, our work focuses on understanding the perceptual differences between human and machine decision-making.

\subsection{Human-centric evaluations}
Neuroscientists have studied the neural encoding process for visual recognition by examining the relationship between primate neuron information and neural network (NN) latent representations \cite{afraz2014neural}. Yamins \textit{et al.} compared human visual cortex signals with NN latent features on classification tasks \cite{yamins2014performance}, while Schrimpf \textit{et al.} introduced Brain-score to quantify the similarity between brain neurons and machine learning (ML) models \cite{schrimpf2020brain}. Rajalingham \textit{et al.} showed that NNs and humans/monkeys have different visual perceptions \cite{rajalingham2018large}. However, this class of work focuses on understanding information at the neuron level rather than on task outcomes.

Machine learning research has also proposed human-centric evaluation frameworks that explain the machine decision-making process to human users. While early studies focused on models trained on tabular data~\cite{lage2019human,poursabzi2021manipulating}, recent work has expanded the scope to more complex vision tasks. Some studies investigate the important segments \cite{zhang2019dissonance} or pixels \cite{biessmann2019psychophysics} of images that contribute to the final perception. Others show how latent features can help humans understand neural networks \cite{borowski2021exemplary, zimmermann2021well}. Some try to find the best way to explain machine predictions \cite{nguyen2021effectiveness,kim2022hive}. Other machine learning research collects human guessing distributions via multiple judgments per image in the classification datasets. This enables researchers to select the best neural network and cognitive models that replicate human performance \cite{battleday2020capturing}, and to make the best use of human disagreements during training \cite{peterson2019human,fornaciari-etal-2021-beyond}. All of this work attempts to align human and machine perceptions, either by providing better explanations or by building more human-like neural models. In contrast, our focus is on understanding the perceptual differences between humans and machines so that we can seek better complimentary teaming between the two.

\subsection{Human-AI collaboration}
Numerous studies have investigated how to effectively leverage human intelligence to annotate the best data set for AI model training. Sheng et al. studied the trade-off between label quality, quantity, and the number of votes per sample given a limited budget \cite{sheng2008get}, while Nguyen et al. demonstrated how to utilize annotators with varying levels of domain expertise \cite{nguyen2015combining}. 

Active learning (AL) methods collect batches of human annotations and iteratively fine-tune models~\cite{han2021iterative}. AL has been applied to a wide range of vision tasks, such as image classification \cite{joshi2009multi, ranganathan2017deep} and segmentation \cite{yao2012interactive}, resulting in improved explainability and more trustworthy predictions. AL has also been used in object detection \cite{roy2018deep,madono2020efficient} and segmentation \cite{yang2017suggestive,ravanbakhsh2020human} to efficiently detect occluded or tiny objects, as well as in image restoration \cite{weber2020draw,roels2019human} to alleviate overfitting in pure learning-based methods and add semantic filling to restore the image. Crowdsourcing methods also utilize annotators to modify class definitions, reduce ambiguity, and include outlier cases \cite{chang2017revolt,kulesza2014structured}. Human in the Loop (HITL) surveys \cite{wu2021survey} and AL surveys \cite{ren2021survey} have also been conducted.


In addition to using human labor as annotators, studies have explored different forms of human-AI collaboration. InstructGPT \cite{ouyang2022training} collects human preferences in dialog systems and uses reinforcement learning (RL), providing human feedback to Large Language Models. A similar RL approach has been applied to blind spot detection \cite{ramakrishnan2018discovering, ramakrishnan2019overcoming}. Human workers have been asked to diagnose AI system failures given semantic explanations of the AI decision \cite{nushi2018towards,nushi2017human}. AI systems also serve as assistants to human decision makers~\cite{bansal2019updates,srivastava2020empirical,bansal2021does}.

 We share a similar framing as many human-AI collaboration researchers, however, our study seeks to understand the expertise difference between humans and machines.


\section{Hypothesis testing}

\subsection{Hypothesis test-1: comparisons of human-machine overall perceptual differences}
\label{Htest:1}

We statistically test whether the perceptual differences between humans and machines are significant or not, through the comparisons of confusion matrices. For human annotations given by CIFAR-10N, we denote the three confusion matrices in Figure \ref{Evidence 101}(a, b, c) as $\mathbf{C}^{\text{HH}}, \mathbf{C}^{\text{MM}}$ and $\mathbf{C}^{\text{HM}}$, respectively. Note that the diagonal entries and off-diagonal entries are supposed to have different distributions. We organize the hypothesis testing w.r.t. either part as follows. 

For $\text{T}_1, \text{T}_2 \in \{\text{HH}, \text{MM}, \text{HM}\}$, the null hypothesis of test-1 $\mathbf{H_0}$ and the corresponding alternate hypothesis $\mathbf{H_1}$ are then defined as:
\begin{align*}
    &\mathbf{H_0} \text{ (Diagonal)}: \text{ There are no significant differences between }\mathbf{C}^{\text{T}_1}_{p,p} \text{ and }\mathbf{C}^{\text{T}_2}_{p,p}, \text{for }p\in[10], \text{T}_1\neq \text{T}_2;\\
    &\mathbf{H_1}\text{ (Diagonal)}: \text{ There are significant differences between }\mathbf{C}^{\text{T}_1}_{p,p} \text{ and }\mathbf{C}^{\text{T}_2}_{p,p}, \text{for }p\in[10], \text{T}_1\neq \text{T}_2.
\end{align*}
\begin{align*}
    &\mathbf{H_0} \text{ (Off-diagonal)}: \text{ There are no significant differences between }\mathbf{C}^{\text{T}_1}_{p,q} \text{ and }\mathbf{C}^{\text{T}_2}_{p,q}, \text{for }p, q\in[10], p\neq q, \text{T}_1\neq \text{T}_2;\\
    &\mathbf{H_1}\text{ (Off-diagonal)}: \text{ There are significant differences between }\mathbf{C}^{\text{T}_1}_{p,q} \text{ and }\mathbf{C}^{\text{T}_2}_{p,q}, \text{for }p, q\in[10],  p\neq q, \text{T}_1\neq \text{T}_2.
\end{align*}
As specified in the above formulation, we adopt the element-wise difference between (1) $\mathbf{C}^{\text{MM}}$ and $\mathbf{C}^{\text{HH}}$; (2) $\mathbf{C}^{\text{HM}}$ and $\mathbf{C}^{\text{HH}}$; (3) $\mathbf{C}^{\text{MM}}$ and $\mathbf{C}^{\text{HM}}$; to quantify the statistical differences between machine and human perception in the confusion matrices. Table \ref{tab:test1} summarizes the hypothesis testing results. For both diagonal and off-diagonal tests, we observe that the $p$-values for (1) MM V.S. HH and (3) MM V.S. HM satisfy that $p\leq \alpha=0.05$, indicating that we should reject the null hypothesis $\text{H}_0$. Thus, we accept the alternate hypothesis $\text{H}_1$ and conclude that there exist significant differences between $\mathbf{C}^{\text{T}_1}_{p,q} \text{ and }\mathbf{C}^{\text{T}_2}_{p,q}, \text{for }p, q\in[10], \text{T}_1\neq \text{T}_2.$

\input{appendix/fig_tex/tab_htest1_overall}

\subsection{Hypothesis test2: Correlation of accuracy to the machine confidence score}
\label{Htest:2}
We statistically test whether the perceptual differences between the human and the machine classifier are significant or not, through the performances on either machine confidence (Figure~\ref{Evidence 102} (a)) or machine agreement (Figure~\ref{Evidence 102} (b)). Specifically, we adopt the Ordinary Least Squares function to fit the individual points that appeared in Figure~\ref{Evidence 102} for machine accuracy and human accuracy. Mathematically, we have: $Y=\beta_0+\beta_1\cdot X,$ where $Y=\text{Machine/Human Accuracy}$, $\beta_0$ is the off-set and $\beta_1$ is a constant denoting the slope of the fitted line. When $X=\text{Machine confidence}$, we have:
\begin{align}\label{eq:test2}
    \text{(Human)}\quad Y=0.1099X+0.7010; \qquad \text{(Machine)}\quad Y=1.0000X+4.077e^{-16}.
\end{align} 
The $p$-value for all coefficients appearing in Eqn. (\ref{eq:test2}) satisfies that $p<e^{-3}\leq \alpha=0.05$, indicating that we should reject the null hypothesis. In other words, the variable $X$ has statistically significant effects on $Y$. What is more, the human performance yields a much smaller $\beta_1$, i.e., $0.1099<1.0000$. Thus, the difficulty level for machine learning models (metric: machine confidence) has less impact on human performance than on machine learning models.

\subsection{Hypothesis test3: Correlation of accuracy to the machine agreement}
\label{Htest:3}

Proceeding Hypothesis test2, when $X=\text{the number of correct predictions among machines}$, we have:
\begin{align}\label{eq:test3}
    \text{(Human)}\quad Y=0.0836X+0.67811; \qquad \text{(Machine)}\quad Y=1.0000X+9.021e^{-17}.
\end{align} 
The $p$-value for all coefficients appearing in Eqn. \ref{eq:test2} satisfies that $p<0.004\leq \alpha=0.05$, indicating that we should reject the null hypothesis. In other words, the variable $X$ has statistically significant effects on $Y$. What is more, human performance yields a much smaller $\beta_1$, i.e., $0.0836<1.0000$. Thus, the difficulty level for machine learning models (metric: number of correct predictions made by machine learning models) has less impact on human performance than on machine learning models.

\subsection{Hypothesis test4: Picking the threshold for human-machine collaboration} 
\label{Htest:swap}

We propose a realistic mode of human-machine collaboration by replacing unconfident model predictions with human annotations. Given a confidence threshold $\eta$, we denote by $\text{Acc}_{\text{col, ml}}(\eta)$ as the test accuracy of replacing unconfident model predictions trained with a specific method by human annotations. We adopt paired t-test to see: for all robust methods, whether replacing unconfident model predictions (w.r.t. a fixed $\eta^*$) by human annotations could be consistently the best among such a human-machine collaboration (i.e., $\text{Acc}^*_{\text{col, ml}}$). Mathematically, 
\begin{align*}
    &\mathbf{H_0}: \text{ There are no significant differences between }\text{Acc}_{\text{col, ml}}(\eta^*) \text{ and }\text{Acc}^*_{\text{col, ml}}, \text{ for a list of robust methods};\\
    &\mathbf{H_1}: \text{ There are significant differences between }\text{Acc}_{\text{col, ml}}(\eta^*) \text{ and }\text{Acc}^*_{\text{col, ml}}, \text{ for a list of robust methods}.
\end{align*}
The hypothesis testing results are summarized in Table~\ref{tab:test_collabrate}. We are interested in two lines of methods: methods require a pre-trained model or multiple networks. For the pre-trained model, we observe that the $p$-value for $\eta=0.6$ satisfies that $p\geq \alpha=0.05$, indicating that we should accept the null hypothesis $\text{H}_0$, and conclude that there exists no significant differences between $\text{Acc}_{\text{col, ml}}(\eta^*=0.6) \text{ and }\text{Acc}^*_{\text{col, ml}}.$

\input{appendix/fig_tex/tab_htest4_swap}

\section{Additional results}

\subsection{Training vision models on different labels or vision backbones}

In section \ref{sec:more training scheme}, we conduct an investigation to support our finding that machines are more similar to other machines and that humans have different statistics. Figure~\ref{Evidence 106} in the main text shows the result of training machines with different sets of training labels. Here in Figure \ref{Evidence 107} we provide the result of training machines with different vision backbones and similarly show that it does not change our findings.

\label{Appendix: various training setup}
\input{appendix/fig_tex/both_incorrect_prec_report_backbone}

\subsection{Human-machine collaboration full report}


In section \ref{sec:collaborate}, we investigate human-machine collaboration.  Figure~\ref{Evidence 202} shows a full report of all possible pairs of human and machine collaborators.  Figure~\ref{Evidence 201} in the main text is a summary of this full report, showing only the average human-machine collaboration and the best machine-machine collaboration in each case.

\input{appendix/fig_tex/Post_hoc_teaming}

%% file: appendix/fig_tex/tab_htest1_overall.tex
\begin{table*}[!h]
\centering
\scalebox{1.0}{
    \begin{tabular}{c|cc|cc|cc}
     \hline
    Diagonal tests &\multicolumn{2}{c}{(1) $\mathbf{C}^{\text{MM}}$ v.s. $\mathbf{C}^{\text{HH}}$}& \multicolumn{2}{c}{(2) $\mathbf{C}^{\text{HM}}$ v.s. $\mathbf{C}^{\text{HH}}$} & \multicolumn{2}{c}{(3) $\mathbf{C}^{\text{MM}}$ v.s. $\mathbf{C}^{\text{HM}}$}  \\
    \hline
    &  Statistic&  $p$-value & Statistic&  $p$-value & Statistic&  $p$-value\\ 
    & 12.57  & 2.40e-10  &  0.04 & 0.97 & 11.56  & 9.14e-10 \\
     \hline \hline
    Off-diagonal tests &\multicolumn{2}{c}{(1) $\mathbf{C}^{\text{MM}}$ v.s. $\mathbf{C}^{\text{HH}}$}& \multicolumn{2}{c}{(2) $\mathbf{C}^{\text{HM}}$ v.s. $\mathbf{C}^{\text{HH}}$} & \multicolumn{2}{c}{(3) $\mathbf{C}^{\text{MM}}$ v.s. $\mathbf{C}^{\text{HM}}$}  \\
    \hline
    &  Statistic&  $p$-value & Statistic&  $p$-value & Statistic&  $p$-value\\ 
    & -12.37   & 9.22e-26  &  -0.04 &  0.97&  -10.44 & 3.40e-20  \\
     \hline
    \end{tabular}}
\caption{Results of hypothesis test-1 with paired student t-test: we adopt the significance value $\alpha=0.05$.}
\label{tab:test1}
\end{table*}

%% file: appendix/fig_tex/tab_htest4_swap.tex
\begin{table}[!htb]
\centering{\scalebox{1.0}{
\begin{tabular}{c|cc|cc|cc}
 \hline
Pre-trained& \multicolumn{2}{c}{ $\text{Acc}_{\text{col, ml}}(0.4) \text{ v.s. }\text{Acc}^*_{\text{col, ml}}$}& \multicolumn{2}{c}{ $\text{Acc}_{\text{col, ml}}(0.5) \text{ v.s. }\text{Acc}^*_{\text{col, ml}}$ } & \multicolumn{2}{c}{ $\text{Acc}_{\text{col, ml}}(0.6) \text{ v.s. }\text{Acc}^*_{\text{col, ml}}$}  \\
\hline
& Statistic&  $p$-value & Statistic&  $p$-value & Statistic&  $p$-value\\ 
&-18.96  & 7.51e-6  &  -7.95 & 0.0005 & -2.49  & 0.0555 \\
 \hline\hline
Multi-nets& \multicolumn{2}{c}{ $\text{Acc}_{\text{col, ml}}(0.4) \text{ v.s. }\text{Acc}^*_{\text{col, ml}}$}& \multicolumn{2}{c}{ $\text{Acc}_{\text{col, ml}}(0.5) \text{ v.s. }\text{Acc}^*_{\text{col, ml}}$ } & \multicolumn{2}{c}{ $\text{Acc}_{\text{col, ml}}(0.6) \text{ v.s. }\text{Acc}^*_{\text{col, ml}}$}  \\
\hline
& Statistic&  $p$-value & Statistic&  $p$-value & Statistic&  $p$-value\\ 
& -66.25 & 0.0002  &  -37.57 & 0.0007 & -34.12  & 0.0009 \\
 \hline
\end{tabular}}}
\caption{Results of hypothesis test2 with paired student t-test: we adopt the significance value $\alpha=0.05$.}
\label{tab:test_collabrate}
\end{table}

%% file: appendix/fig_tex/both_incorrect_prec_report_backbone.tex
\begin{figure*}[htp]
    \centering  
    \includegraphics[width=0.8\textwidth]{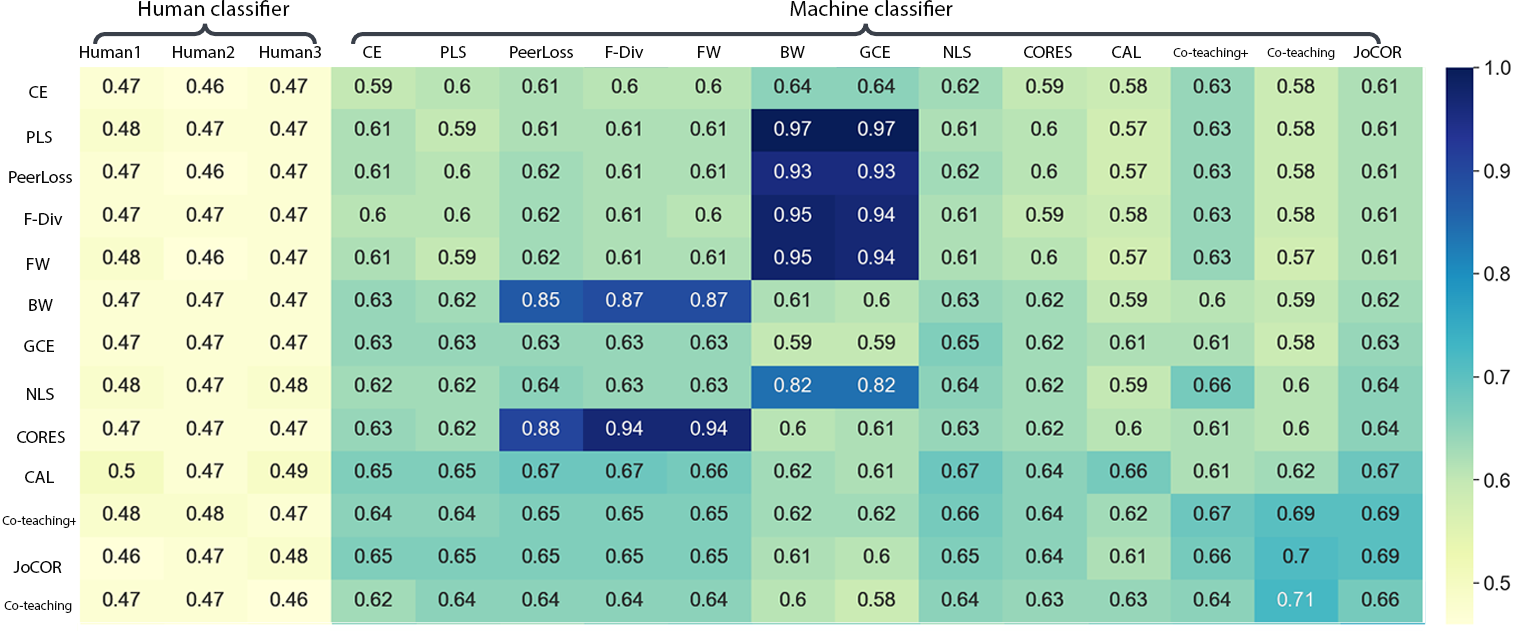}

    \caption{\textbf{Matching Percentage on balanced set:} The figure visualizes the matching percentage between each machine classifier and each other human/machine classifier, on a balanced set. The machine classifiers are not all trained with the same vision backbone(Resnet, VGG, Inception), yet the results show machines tend to make judgements which match other machines more than they match humans. }
    
    
    \label{Evidence 107}
\end{figure*}

%% file: appendix/fig_tex/Post_hoc_teaming.tex
\begin{figure}[htp]
    \begin{center}
    \subfigure[Oracle mode]{
        \includegraphics[width=0.8\textwidth]{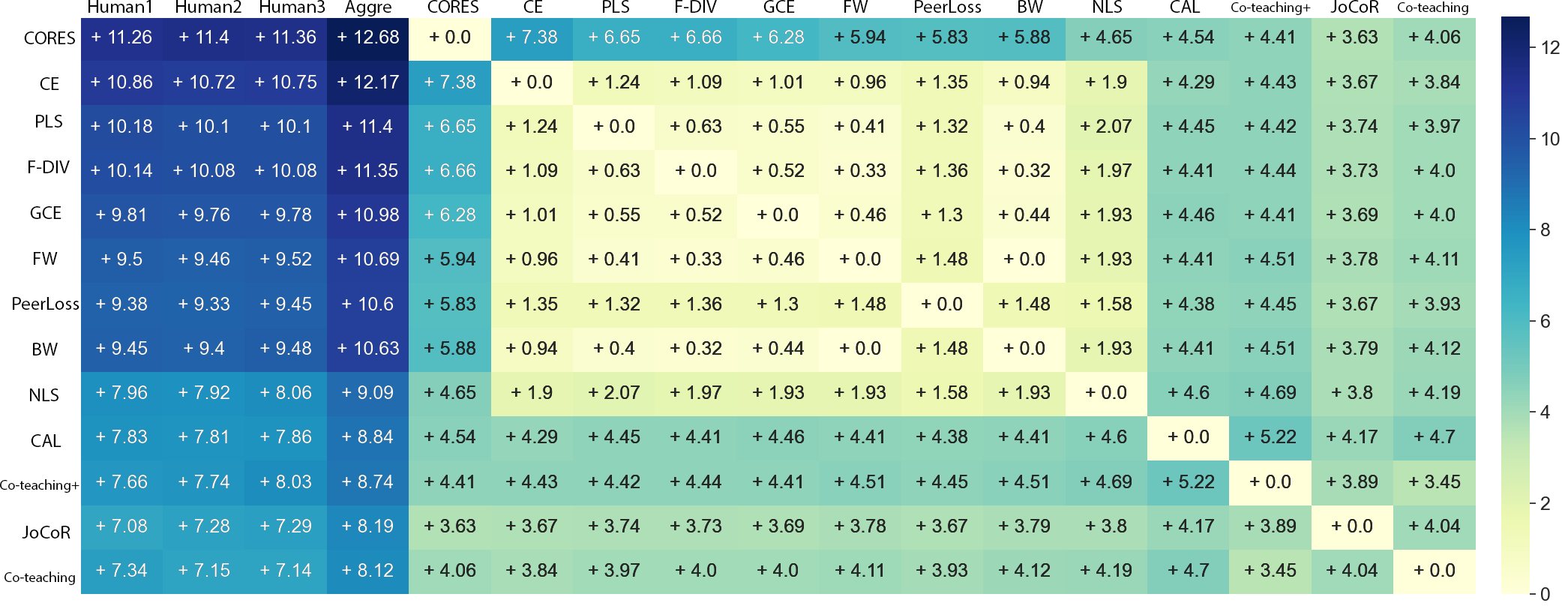}
    }
    \subfigure[Realistic mode]{
        \includegraphics[width=0.8\textwidth]{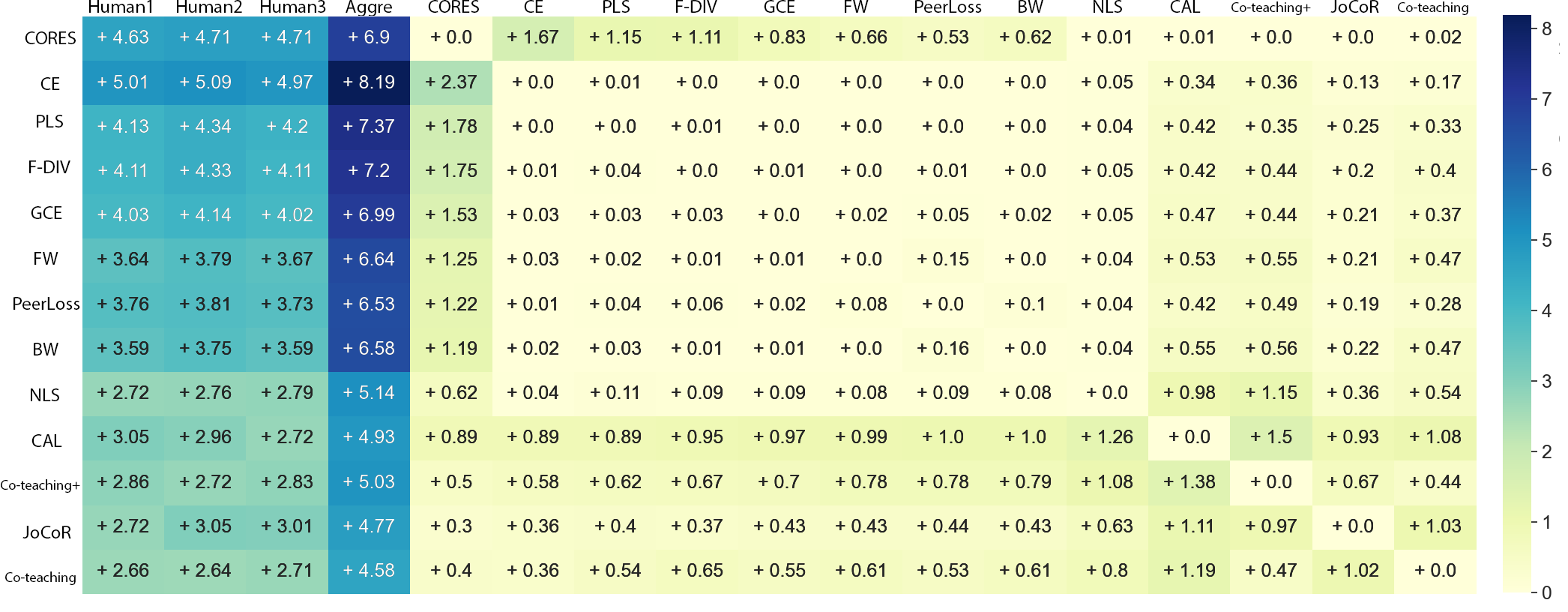}
    }

    \end{center}
    \vspace{-0.7em}
    \caption{\textbf{Post-hoc teaming:} The figure shows original model performance, and the boost from teaming options. We compared all the permutations and visualize the best teaming combinations using a colormap. A darker color indicates a greater boost. \textbf{(a)} Oracle mode is the upper bound from perfect teaming, \textbf{(b)} realistic mode is from a simplistic real algorithm. The results show the value of human-machine complementary teaming. Introducing a low performance human to the teaming system causes more boost than introducing a higher performance machine classifier. }
    \label{Evidence 202}
    \vspace{-1.5em}
\end{figure}

%% file: main_aaai25.bbl
\begin{thebibliography}{99}
\providecommand{\natexlab}[1]{#1}

\bibitem[{Abnar and Zuidema(2020)}]{abnar2020quantifying}
Abnar, S.; and Zuidema, W. 2020.
\newblock Quantifying attention flow in transformers.
\newblock \emph{arXiv preprint arXiv:2005.00928}.

\bibitem[{Afraz, Yamins, and DiCarlo(2014)}]{afraz2014neural}
Afraz, A.; Yamins, D.~L.; and DiCarlo, J.~J. 2014.
\newblock Neural mechanisms underlying visual object recognition.
\newblock In \emph{Cold Spring Harbor symposia on quantitative biology}, volume~79, 99--107. Cold Spring Harbor Laboratory Press.

\bibitem[{Agrawal et~al.(2014)Agrawal, Stansbury, Malik, and Gallant}]{agrawal2014pixels}
Agrawal, P.; Stansbury, D.; Malik, J.; and Gallant, J.~L. 2014.
\newblock Pixels to voxels: modeling visual representation in the human brain.
\newblock \emph{arXiv preprint arXiv:1407.5104}.

\bibitem[{Alicioglu and Sun(2022)}]{alicioglu2022survey}
Alicioglu, G.; and Sun, B. 2022.
\newblock A survey of visual analytics for Explainable Artificial Intelligence methods.
\newblock \emph{Computers \& Graphics}, 102: 502--520.

\bibitem[{Bansal et~al.(2019)Bansal, Nushi, Kamar, Weld, Lasecki, and Horvitz}]{bansal2019updates}
Bansal, G.; Nushi, B.; Kamar, E.; Weld, D.~S.; Lasecki, W.~S.; and Horvitz, E. 2019.
\newblock Updates in {Human}-{AI} teams: Understanding and addressing the performance/compatibility tradeoff.
\newblock In \emph{Proceedings of the AAAI Conference on Artificial Intelligence}, volume~33, 2429--2437.

\bibitem[{Bansal et~al.(2021)Bansal, Wu, Zhou, Fok, Nushi, Kamar, Ribeiro, and Weld}]{bansal2021does}
Bansal, G.; Wu, T.; Zhou, J.; Fok, R.; Nushi, B.; Kamar, E.; Ribeiro, M.~T.; and Weld, D. 2021.
\newblock Does the whole exceed its parts? {The} effect of {AI} explanations on complementary team performance.
\newblock In \emph{Proceedings of the 2021 CHI Conference on Human Factors in Computing Systems}, 1--16.

\bibitem[{Battleday, Peterson, and Griffiths(2020)}]{battleday2020capturing}
Battleday, R.~M.; Peterson, J.~C.; and Griffiths, T.~L. 2020.
\newblock Capturing human categorization of natural images by combining deep networks and cognitive models.
\newblock \emph{Nature communications}, 11(1): 1--14.

\bibitem[{Biessmann and Refiano(2019)}]{biessmann2019psychophysics}
Biessmann, F.; and Refiano, D.~I. 2019.
\newblock A psychophysics approach for quantitative comparison of interpretable computer vision models.
\newblock \emph{arXiv preprint arXiv:1912.05011}.

\bibitem[{Borowski et~al.(2020)Borowski, Zimmermann, Schepers, Geirhos, Wallis, Bethge, and Brendel}]{borowski2021exemplary}
Borowski, J.; Zimmermann, R.~S.; Schepers, J.; Geirhos, R.; Wallis, T.~S.; Bethge, M.; and Brendel, W. 2020.
\newblock Exemplary natural images explain {CNN} activations better than state-of-the-art feature visualization.
\newblock \emph{arXiv preprint arXiv:2010.12606}.

\bibitem[{Branson et~al.(2010)Branson, Wah, Schroff, Babenko, Welinder, Perona, and Belongie}]{branson2010visual}
Branson, S.; Wah, C.; Schroff, F.; Babenko, B.; Welinder, P.; Perona, P.; and Belongie, S. 2010.
\newblock Visual recognition with humans in the loop.
\newblock In \emph{Computer Vision--ECCV 2010: 11th European Conference on Computer Vision, Heraklion, Crete, Greece, September 5-11, 2010, Proceedings, Part IV 11}, 438--451. Springer.

\bibitem[{Chang, Amershi, and Kamar(2017)}]{chang2017revolt}
Chang, J.~C.; Amershi, S.; and Kamar, E. 2017.
\newblock Revolt: Collaborative crowdsourcing for labeling machine learning datasets.
\newblock In \emph{Proceedings of the 2017 CHI Conference on Human Factors in Computing Systems}, 2334--2346.

\bibitem[{Chefer, Gur, and Wolf(2021)}]{chefer2021transformer}
Chefer, H.; Gur, S.; and Wolf, L. 2021.
\newblock Transformer interpretability beyond attention visualization.
\newblock In \emph{Proceedings of the IEEE/CVF Conference on Computer Vision and Pattern Recognition}, 782--791.

\bibitem[{Chen et~al.(2023{\natexlab{a}})Chen, Shah, Wang, Tao, Wang, Xie, Sugiyama, Singh, and Raj}]{chen2023imprecise}
Chen, H.; Shah, A.; Wang, J.; Tao, R.; Wang, Y.; Xie, X.; Sugiyama, M.; Singh, R.; and Raj, B. 2023{\natexlab{a}}.
\newblock Imprecise label learning: A unified framework for learning with various imprecise label configurations.
\newblock \emph{arXiv preprint arXiv:2305.12715}.

\bibitem[{Chen et~al.(2023{\natexlab{b}})Chen, Wang, Shah, Tao, Wei, Xie, Sugiyama, and Raj}]{chen2023understanding}
Chen, H.; Wang, J.; Shah, A.; Tao, R.; Wei, H.; Xie, X.; Sugiyama, M.; and Raj, B. 2023{\natexlab{b}}.
\newblock Understanding and mitigating the label noise in pre-training on downstream tasks.
\newblock \emph{arXiv preprint arXiv:2309.17002}.

\bibitem[{Cheng et~al.(2020)Cheng, Zhu, Li, Gong, Sun, and Liu}]{cheng2020learning}
Cheng, H.; Zhu, Z.; Li, X.; Gong, Y.; Sun, X.; and Liu, Y. 2020.
\newblock Learning with instance-dependent label noise: A sample sieve approach.
\newblock \emph{arXiv preprint arXiv:2010.02347}.

\bibitem[{Cheng et~al.(2021)Cheng, Zhu, Li, Gong, Sun, and Liu}]{cheng2021learning}
Cheng, H.; Zhu, Z.; Li, X.; Gong, Y.; Sun, X.; and Liu, Y. 2021.
\newblock Learning with Instance-Dependent Label Noise: A Sample Sieve Approach.
\newblock In \emph{International Conference on Learning Representations}.

\bibitem[{Dabkowski and Gal(2017)}]{dabkowski2017real}
Dabkowski, P.; and Gal, Y. 2017.
\newblock Real time image saliency for black box classifiers.
\newblock \emph{Advances in neural information processing systems}, 30.

\bibitem[{Deng et~al.(2009)Deng, Dong, Socher, Li, Li, and Fei-Fei}]{deng2009imagenet}
Deng, J.; Dong, W.; Socher, R.; Li, L.-J.; Li, K.; and Fei-Fei, L. 2009.
\newblock Imagenet: A large-scale hierarchical image database.
\newblock In \emph{2009 IEEE conference on computer vision and pattern recognition}, 248--255. Ieee.

\bibitem[{Dosovitskiy et~al.(2020)Dosovitskiy, Beyer, Kolesnikov, Weissenborn, Zhai, Unterthiner, Dehghani, Minderer, Heigold, Gelly et~al.}]{dosovitskiy2020image}
Dosovitskiy, A.; Beyer, L.; Kolesnikov, A.; Weissenborn, D.; Zhai, X.; Unterthiner, T.; Dehghani, M.; Minderer, M.; Heigold, G.; Gelly, S.; et~al. 2020.
\newblock An image is worth 16x16 words: Transformers for image recognition at scale.
\newblock \emph{arXiv preprint arXiv:2010.11929}.

\bibitem[{Erhan et~al.(2009)Erhan, Bengio, Courville, and Vincent}]{erhan2009visualizing}
Erhan, D.; Bengio, Y.; Courville, A.; and Vincent, P. 2009.
\newblock Visualizing higher-layer features of a deep network.
\newblock \emph{University of Montreal}, 1341(3): 1.

\bibitem[{Fornaciari et~al.(2021)Fornaciari, Uma, Paun, Plank, Hovy, and Poesio}]{fornaciari-etal-2021-beyond}
Fornaciari, T.; Uma, A.; Paun, S.; Plank, B.; Hovy, D.; and Poesio, M. 2021.
\newblock Beyond Black {\&} White: Leveraging Annotator Disagreement via Soft-Label Multi-Task Learning.
\newblock In \emph{Proceedings of the 2021 Conference of the North American Chapter of the Association for Computational Linguistics: Human Language Technologies}, 2591--2597. Online: Association for Computational Linguistics.

\bibitem[{Guo et~al.(2021)Guo, Liu, Mu, and Hu}]{guo2021beyond}
Guo, M.-H.; Liu, Z.-N.; Mu, T.-J.; and Hu, S.-M. 2021.
\newblock Beyond self-attention: External attention using two linear layers for visual tasks.
\newblock \emph{arXiv preprint arXiv:2105.02358}.

\bibitem[{Han et~al.(2018)Han, Yao, Yu, Niu, Xu, Hu, Tsang, and Sugiyama}]{han2018co}
Han, B.; Yao, Q.; Yu, X.; Niu, G.; Xu, M.; Hu, W.; Tsang, I.; and Sugiyama, M. 2018.
\newblock Co-teaching: Robust training of deep neural networks with extremely noisy labels.
\newblock \emph{Advances in neural information processing systems}, 31.

\bibitem[{Han, Dong, and Demartini(2021)}]{han2021iterative}
Han, L.; Dong, X.; and Demartini, G. 2021.
\newblock Iterative Human-in-the-Loop Discovery of Unknown Unknowns in Image Datasets.
\newblock In \emph{Proceedings of the AAAI Conference on Human Computation and Crowdsourcing}, volume~9, 72--83.

\bibitem[{He et~al.(2016)He, Zhang, Ren, and Sun}]{he2016deep}
He, K.; Zhang, X.; Ren, S.; and Sun, J. 2016.
\newblock Deep residual learning for image recognition.
\newblock In \emph{Proceedings of the IEEE conference on computer vision and pattern recognition}, 770--778.

\bibitem[{Huang et~al.(2023)Huang, Kang, Liu, Salvado, Rakotoarivelo, Wang, and Liu}]{huang2023paddles}
Huang, H.; Kang, H.; Liu, S.; Salvado, O.; Rakotoarivelo, T.; Wang, D.; and Liu, T. 2023.
\newblock Paddles: Phase-amplitude spectrum disentangled early stopping for learning with noisy labels.
\newblock In \emph{Proceedings of the IEEE/CVF International Conference on Computer Vision}, 16719--16730.

\bibitem[{Iwana, Kuroki, and Uchida(2019)}]{iwana2019explaining}
Iwana, B.~K.; Kuroki, R.; and Uchida, S. 2019.
\newblock Explaining convolutional neural networks using softmax gradient layer-wise relevance propagation.
\newblock In \emph{2019 IEEE/CVF International Conference on Computer Vision Workshop (ICCVW)}, 4176--4185. IEEE.

\bibitem[{Joshi, Porikli, and Papanikolopoulos(2009)}]{joshi2009multi}
Joshi, A.~J.; Porikli, F.; and Papanikolopoulos, N. 2009.
\newblock Multi-class active learning for image classification.
\newblock In \emph{2009 ieee conference on computer vision and pattern recognition}, 2372--2379. IEEE.

\bibitem[{Kim et~al.(2022)Kim, Meister, Ramaswamy, Fong, and Russakovsky}]{kim2022hive}
Kim, S.~S.; Meister, N.; Ramaswamy, V.~V.; Fong, R.; and Russakovsky, O. 2022.
\newblock Hive: evaluating the human interpretability of visual explanations.
\newblock In \emph{Computer Vision--ECCV 2022: 17th European Conference, Tel Aviv, Israel, October 23--27, 2022, Proceedings, Part XII}, 280--298. Springer.

\bibitem[{Kovashka et~al.(2016)Kovashka, Russakovsky, Fei-Fei, Grauman et~al.}]{kovashka2016crowdsourcing}
Kovashka, A.; Russakovsky, O.; Fei-Fei, L.; Grauman, K.; et~al. 2016.
\newblock Crowdsourcing in computer vision.
\newblock \emph{Foundations and Trends{\textregistered} in computer graphics and Vision}, 10(3): 177--243.

\bibitem[{Krizhevsky, Hinton et~al.(2009)}]{krizhevsky2009learning}
Krizhevsky, A.; Hinton, G.; et~al. 2009.
\newblock Learning multiple layers of features from tiny images.
\newblock Technical report, Citeseer.

\bibitem[{Krizhevsky, Sutskever, and Hinton(2012)}]{krizhevsky2012imagenet}
Krizhevsky, A.; Sutskever, I.; and Hinton, G.~E. 2012.
\newblock Imagenet classification with deep convolutional neural networks.
\newblock \emph{Advances in neural information processing systems}, 25.

\bibitem[{Kulesza et~al.(2014)Kulesza, Amershi, Caruana, Fisher, and Charles}]{kulesza2014structured}
Kulesza, T.; Amershi, S.; Caruana, R.; Fisher, D.; and Charles, D. 2014.
\newblock Structured labeling for facilitating concept evolution in machine learning.
\newblock In \emph{Proceedings of the SIGCHI Conference on Human Factors in Computing Systems}, 3075--3084.

\bibitem[{Kuzovkin et~al.(2018)Kuzovkin, Vicente, Petton, Lachaux, Baciu, Kahane, Rheims, Vidal, and Aru}]{kuzovkin2018activations}
Kuzovkin, I.; Vicente, R.; Petton, M.; Lachaux, J.-P.; Baciu, M.; Kahane, P.; Rheims, S.; Vidal, J.~R.; and Aru, J. 2018.
\newblock Activations of deep convolutional neural networks are aligned with gamma band activity of human visual cortex.
\newblock \emph{Communications biology}, 1(1): 1--12.

\bibitem[{Lage et~al.(2019)Lage, Chen, He, Narayanan, Kim, Gershman, and Doshi-Velez}]{lage2019human}
Lage, I.; Chen, E.; He, J.; Narayanan, M.; Kim, B.; Gershman, S.~J.; and Doshi-Velez, F. 2019.
\newblock Human evaluation of models built for interpretability.
\newblock In \emph{Proceedings of the AAAI Conference on Human Computation and Crowdsourcing}, volume~7, 59--67.

\bibitem[{LeCun et~al.(1989)LeCun, Boser, Denker, Henderson, Howard, Hubbard, and Jackel}]{lecun1989backpropagation}
LeCun, Y.; Boser, B.; Denker, J.~S.; Henderson, D.; Howard, R.~E.; Hubbard, W.; and Jackel, L.~D. 1989.
\newblock Backpropagation applied to handwritten zip code recognition.
\newblock \emph{Neural computation}, 1(4): 541--551.

\bibitem[{Liu et~al.(2020)Liu, Niles-Weed, Razavian, and Fernandez-Granda}]{liu2020early}
Liu, S.; Niles-Weed, J.; Razavian, N.; and Fernandez-Granda, C. 2020.
\newblock Early-learning regularization prevents memorization of noisy labels.
\newblock \emph{Advances in neural information processing systems}, 33: 20331--20342.

\bibitem[{Liu et~al.(2022)Liu, Zhu, Qu, and You}]{liu2022robust}
Liu, S.; Zhu, Z.; Qu, Q.; and You, C. 2022.
\newblock Robust training under label noise by over-parameterization.
\newblock In \emph{International Conference on Machine Learning}, 14153--14172. PMLR.

\bibitem[{Liu and Tao(2015)}]{liu2015classification}
Liu, T.; and Tao, D. 2015.
\newblock Classification with noisy labels by importance reweighting.
\newblock \emph{IEEE Transactions on pattern analysis and machine intelligence}, 38(3): 447--461.

\bibitem[{Liu and Guo(2020)}]{liu2020peer}
Liu, Y.; and Guo, H. 2020.
\newblock Peer loss functions: Learning from noisy labels without knowing noise rates.
\newblock In \emph{International conference on machine learning}, 6226--6236. PMLR.

\bibitem[{Lukasik et~al.(2020)Lukasik, Bhojanapalli, Menon, and Kumar}]{lukasik2020does}
Lukasik, M.; Bhojanapalli, S.; Menon, A.; and Kumar, S. 2020.
\newblock Does label smoothing mitigate label noise?
\newblock In \emph{International Conference on Machine Learning}, 6448--6458. PMLR.

\bibitem[{Madono et~al.(2020)Madono, Nakano, Kobayashi, and Ogawa}]{madono2020efficient}
Madono, K.; Nakano, T.; Kobayashi, T.; and Ogawa, T. 2020.
\newblock Efficient Human-In-The-Loop Object Detection using Bi-Directional Deep SORT and Annotation-Free Segment Identification.
\newblock In \emph{2020 Asia-Pacific Signal and Information Processing Association Annual Summit and Conference (APSIPA ASC)}, 1226--1233. IEEE.

\bibitem[{Montavon et~al.(2017)Montavon, Lapuschkin, Binder, Samek, and M{\"u}ller}]{montavon2017explaining}
Montavon, G.; Lapuschkin, S.; Binder, A.; Samek, W.; and M{\"u}ller, K.-R. 2017.
\newblock Explaining nonlinear classification decisions with deep taylor decomposition.
\newblock \emph{Pattern recognition}, 65: 211--222.

\bibitem[{Mur et~al.(2013)Mur, Meys, Bodurka, Goebel, Bandettini, and Kriegeskorte}]{mur2013human}
Mur, M.; Meys, M.; Bodurka, J.; Goebel, R.; Bandettini, P.~A.; and Kriegeskorte, N. 2013.
\newblock Human object-similarity judgments reflect and transcend the primate-IT object representation.
\newblock \emph{Frontiers in psychology}, 4: 128.

\bibitem[{Natarajan et~al.(2013)Natarajan, Dhillon, Ravikumar, and Tewari}]{natarajan2013learning}
Natarajan, N.; Dhillon, I.~S.; Ravikumar, P.~K.; and Tewari, A. 2013.
\newblock Learning with noisy labels.
\newblock \emph{Advances in neural information processing systems}, 26.

\bibitem[{Nguyen, Wallace, and Lease(2015)}]{nguyen2015combining}
Nguyen, A.~T.; Wallace, B.~C.; and Lease, M. 2015.
\newblock Combining crowd and expert labels using decision theoretic active learning.
\newblock In \emph{Third AAAI conference on human computation and crowdsourcing}.

\bibitem[{Nguyen, Kim, and Nguyen(2021)}]{nguyen2021effectiveness}
Nguyen, G.; Kim, D.; and Nguyen, A. 2021.
\newblock The effectiveness of feature attribution methods and its correlation with automatic evaluation scores.
\newblock \emph{Advances in Neural Information Processing Systems}, 34: 26422--26436.

\bibitem[{Nushi, Kamar, and Horvitz(2018)}]{nushi2018towards}
Nushi, B.; Kamar, E.; and Horvitz, E. 2018.
\newblock Towards accountable {AI}: Hybrid human-machine analyses for characterizing system failure.
\newblock In \emph{Proceedings of the AAAI Conference on Human Computation and Crowdsourcing}, volume~6, 126--135.

\bibitem[{Nushi et~al.(2017)Nushi, Kamar, Horvitz, and Kossmann}]{nushi2017human}
Nushi, B.; Kamar, E.; Horvitz, E.; and Kossmann, D. 2017.
\newblock On human intellect and machine failures: Troubleshooting integrative machine learning systems.
\newblock In \emph{Proceedings of the AAAI Conference on Artificial Intelligence}, volume~31.

\bibitem[{Ortiz-Jimenez et~al.(2023)Ortiz-Jimenez, Collier, Nawalgaria, D’Amour, Berent, Jenatton, and Kokiopoulou}]{ortiz2023does}
Ortiz-Jimenez, G.; Collier, M.; Nawalgaria, A.; D’Amour, A.~N.; Berent, J.; Jenatton, R.; and Kokiopoulou, E. 2023.
\newblock When does privileged information explain away label noise?
\newblock In \emph{International Conference on Machine Learning}, 26646--26669. PMLR.

\bibitem[{Ouyang et~al.(2022)Ouyang, Wu, Jiang, Almeida, Wainwright, Mishkin, Zhang, Agarwal, Slama, Ray et~al.}]{ouyang2022training}
Ouyang, L.; Wu, J.; Jiang, X.; Almeida, D.; Wainwright, C.; Mishkin, P.; Zhang, C.; Agarwal, S.; Slama, K.; Ray, A.; et~al. 2022.
\newblock Training language models to follow instructions with human feedback.
\newblock \emph{Advances in Neural Information Processing Systems}, 35: 27730--27744.

\bibitem[{Park et~al.(2024)Park, Choi, Kim, Song, and Lee}]{park2024robust}
Park, D.; Choi, S.; Kim, D.; Song, H.; and Lee, J.-G. 2024.
\newblock Robust data pruning under label noise via maximizing re-labeling accuracy.
\newblock \emph{Advances in Neural Information Processing Systems}, 36.

\bibitem[{Patrini et~al.(2017)Patrini, Rozza, Krishna~Menon, Nock, and Qu}]{patrini2017making}
Patrini, G.; Rozza, A.; Krishna~Menon, A.; Nock, R.; and Qu, L. 2017.
\newblock Making deep neural networks robust to label noise: A loss correction approach.
\newblock In \emph{Proceedings of the IEEE conference on computer vision and pattern recognition}, 1944--1952.

\bibitem[{Peterson et~al.(2019)Peterson, Battleday, Griffiths, and Russakovsky}]{peterson2019human}
Peterson, J.~C.; Battleday, R.~M.; Griffiths, T.~L.; and Russakovsky, O. 2019.
\newblock Human uncertainty makes classification more robust.
\newblock In \emph{Proceedings of the IEEE/CVF International Conference on Computer Vision}, 9617--9626.

\bibitem[{Poursabzi-Sangdeh et~al.(2021)Poursabzi-Sangdeh, Goldstein, Hofman, Wortman~Vaughan, and Wallach}]{poursabzi2021manipulating}
Poursabzi-Sangdeh, F.; Goldstein, D.~G.; Hofman, J.~M.; Wortman~Vaughan, J.~W.; and Wallach, H. 2021.
\newblock Manipulating and measuring model interpretability.
\newblock In \emph{Proceedings of the 2021 CHI conference on human factors in computing systems}, 1--52.

\bibitem[{Rajalingham et~al.(2018)Rajalingham, Issa, Bashivan, Kar, Schmidt, and DiCarlo}]{rajalingham2018large}
Rajalingham, R.; Issa, E.~B.; Bashivan, P.; Kar, K.; Schmidt, K.; and DiCarlo, J.~J. 2018.
\newblock Large-scale, high-resolution comparison of the core visual object recognition behavior of humans, monkeys, and state-of-the-art deep artificial neural networks.
\newblock \emph{Journal of Neuroscience}, 38(33): 7255--7269.

\bibitem[{Ramakrishnan et~al.(2018)Ramakrishnan, Kamar, Dey, Shah, and Horvitz}]{ramakrishnan2018discovering}
Ramakrishnan, R.; Kamar, E.; Dey, D.; Shah, J.; and Horvitz, E. 2018.
\newblock Discovering blind spots in reinforcement learning.
\newblock \emph{arXiv preprint arXiv:1805.08966}.

\bibitem[{Ramakrishnan et~al.(2019)Ramakrishnan, Kamar, Nushi, Dey, Shah, and Horvitz}]{ramakrishnan2019overcoming}
Ramakrishnan, R.; Kamar, E.; Nushi, B.; Dey, D.; Shah, J.; and Horvitz, E. 2019.
\newblock Overcoming blind spots in the real world: Leveraging complementary abilities for joint execution.
\newblock In \emph{Proceedings of the AAAI Conference on Artificial Intelligence}, volume~33, 6137--6145.

\bibitem[{Ranganathan et~al.(2017)Ranganathan, Venkateswara, Chakraborty, and Panchanathan}]{ranganathan2017deep}
Ranganathan, H.; Venkateswara, H.; Chakraborty, S.; and Panchanathan, S. 2017.
\newblock Deep active learning for image classification.
\newblock In \emph{2017 IEEE International Conference on Image Processing (ICIP)}, 3934--3938. IEEE.

\bibitem[{Ravanbakhsh et~al.(2020)Ravanbakhsh, Tschernezki, Last, Klein, Batmanghelich, Tresp, and Nabi}]{ravanbakhsh2020human}
Ravanbakhsh, M.; Tschernezki, V.; Last, F.; Klein, T.; Batmanghelich, K.; Tresp, V.; and Nabi, M. 2020.
\newblock Human-machine collaboration for medical image segmentation.
\newblock In \emph{ICASSP 2020-2020 IEEE International Conference on Acoustics, Speech and Signal Processing (ICASSP)}, 1040--1044. IEEE.

\bibitem[{Ren et~al.(2021)Ren, Xiao, Chang, Huang, Li, Gupta, Chen, and Wang}]{ren2021survey}
Ren, P.; Xiao, Y.; Chang, X.; Huang, P.-Y.; Li, Z.; Gupta, B.~B.; Chen, X.; and Wang, X. 2021.
\newblock A survey of deep active learning.
\newblock \emph{ACM Computing Surveys (CSUR)}, 54(9): 1--40.

\bibitem[{Roels et~al.(2019)Roels, Vernaillen, Kremer, Gon{\c{c}}alves, Aelterman, Luong, Goossens, Philips, Lippens, and Saeys}]{roels2019human}
Roels, J.; Vernaillen, F.; Kremer, A.; Gon{\c{c}}alves, A.; Aelterman, J.; Luong, H.~Q.; Goossens, B.; Philips, W.; Lippens, S.; and Saeys, Y. 2019.
\newblock A “human-in-the-loop” approach for semi-automated image restoration in electron microscopy.
\newblock \emph{bioRxiv}, 644146.

\bibitem[{Roy, Unmesh, and Namboodiri(2018)}]{roy2018deep}
Roy, S.; Unmesh, A.; and Namboodiri, V.~P. 2018.
\newblock Deep active learning for object detection.
\newblock In \emph{BMVC}, 91.

\bibitem[{Schrimpf et~al.(2020)Schrimpf, Kubilius, Hong, Majaj, Rajalingham, Issa, Kar, Bashivan, Prescott-Roy, Geiger et~al.}]{schrimpf2020brain}
Schrimpf, M.; Kubilius, J.; Hong, H.; Majaj, N.~J.; Rajalingham, R.; Issa, E.~B.; Kar, K.; Bashivan, P.; Prescott-Roy, J.; Geiger, F.; et~al. 2020.
\newblock Brain-score: Which artificial neural network for object recognition is most brain-like?
\newblock \emph{BioRxiv}, 407007.

\bibitem[{Selvaraju et~al.(2017)Selvaraju, Cogswell, Das, Vedantam, Parikh, and Batra}]{selvaraju2017grad}
Selvaraju, R.~R.; Cogswell, M.; Das, A.; Vedantam, R.; Parikh, D.; and Batra, D. 2017.
\newblock Grad-cam: Visual explanations from deep networks via gradient-based localization.
\newblock In \emph{Proceedings of the IEEE international conference on computer vision}, 618--626.

\bibitem[{Sheng, Provost, and Ipeirotis(2008)}]{sheng2008get}
Sheng, V.~S.; Provost, F.; and Ipeirotis, P.~G. 2008.
\newblock Get another label? improving data quality and data mining using multiple, noisy labelers.
\newblock In \emph{Proceedings of the 14th ACM SIGKDD international conference on Knowledge discovery and data mining}, 614--622.

\bibitem[{{Shnayder} et~al.(2016){Shnayder}, {Agarwal}, {Frongillo}, and {Parkes}}]{2016arXiv160303151S}
{Shnayder}, V.; {Agarwal}, A.; {Frongillo}, R.; and {Parkes}, D.~C. 2016.
\newblock {Informed Truthfulness in Multi-Task Peer Prediction}.
\newblock \emph{ACM EC}.

\bibitem[{Simonyan and Zisserman(2014)}]{simonyan2014very}
Simonyan, K.; and Zisserman, A. 2014.
\newblock Very deep convolutional networks for large-scale image recognition.
\newblock \emph{arXiv preprint arXiv:1409.1556}.

\bibitem[{Smilkov et~al.(2017)Smilkov, Thorat, Kim, Vi{\'e}gas, and Wattenberg}]{smilkov2017smoothgrad}
Smilkov, D.; Thorat, N.; Kim, B.; Vi{\'e}gas, F.; and Wattenberg, M. 2017.
\newblock Smoothgrad: removing noise by adding noise.
\newblock \emph{arXiv preprint arXiv:1706.03825}.

\bibitem[{Srivastava et~al.(2020)Srivastava, Nushi, Kamar, Shah, and Horvitz}]{srivastava2020empirical}
Srivastava, M.; Nushi, B.; Kamar, E.; Shah, S.; and Horvitz, E. 2020.
\newblock An empirical analysis of backward compatibility in machine learning systems.
\newblock In \emph{Proceedings of the 26th ACM SIGKDD International Conference on Knowledge Discovery \& Data Mining}, 3272--3280.

\bibitem[{Szegedy et~al.(2015)Szegedy, Liu, Jia, Sermanet, Reed, Anguelov, Erhan, Vanhoucke, and Rabinovich}]{szegedy2015going}
Szegedy, C.; Liu, W.; Jia, Y.; Sermanet, P.; Reed, S.; Anguelov, D.; Erhan, D.; Vanhoucke, V.; and Rabinovich, A. 2015.
\newblock Going deeper with convolutions.
\newblock In \emph{Proceedings of the IEEE conference on computer vision and pattern recognition}, 1--9.

\bibitem[{Taleb et~al.(2021)Taleb, Lippert, Klein, and Nabi}]{taleb2021multimodal}
Taleb, A.; Lippert, C.; Klein, T.; and Nabi, M. 2021.
\newblock Multimodal self-supervised learning for medical image analysis.
\newblock In \emph{International Conference on Information Processing in Medical Imaging}, 661--673. Springer.

\bibitem[{Tolstikhin et~al.(2021)Tolstikhin, Houlsby, Kolesnikov, Beyer, Zhai, Unterthiner, Yung, Steiner, Keysers, Uszkoreit et~al.}]{tolstikhin2021mlp}
Tolstikhin, I.~O.; Houlsby, N.; Kolesnikov, A.; Beyer, L.; Zhai, X.; Unterthiner, T.; Yung, J.; Steiner, A.; Keysers, D.; Uszkoreit, J.; et~al. 2021.
\newblock {MLP}-mixer: An all-{MLP} architecture for vision.
\newblock \emph{Advances in Neural Information Processing Systems}, 34.

\bibitem[{Turk(2012)}]{mturk}
Turk, A.~M. 2012.
\newblock Amazon mechanical turk.
\newblock \emph{Retrieved August}, 17: 2012.

\bibitem[{Vaswani et~al.(2017)Vaswani, Shazeer, Parmar, Uszkoreit, Jones, Gomez, Kaiser, and Polosukhin}]{vaswani2017attention}
Vaswani, A.; Shazeer, N.; Parmar, N.; Uszkoreit, J.; Jones, L.; Gomez, A.~N.; Kaiser, {\L}.; and Polosukhin, I. 2017.
\newblock Attention is all you need.
\newblock \emph{Advances in neural information processing systems}, 30.

\bibitem[{Voita et~al.(2019)Voita, Talbot, Moiseev, Sennrich, and Titov}]{voita2019analyzing}
Voita, E.; Talbot, D.; Moiseev, F.; Sennrich, R.; and Titov, I. 2019.
\newblock Analyzing multi-head self-attention: Specialized heads do the heavy lifting, the rest can be pruned.
\newblock \emph{arXiv preprint arXiv:1905.09418}.

\bibitem[{Wang et~al.(2022)Wang, Xiao, Dong, Feng, and Zhao}]{wang2022promix}
Wang, H.; Xiao, R.; Dong, Y.; Feng, L.; and Zhao, J. 2022.
\newblock Promix: Combating label noise via maximizing clean sample utility.
\newblock \emph{arXiv preprint arXiv:2207.10276}.

\bibitem[{Wang et~al.(2024)Wang, Feng, Jiang, Niu, Zhang, and Sugiyama}]{wang2024binary}
Wang, W.; Feng, L.; Jiang, Y.; Niu, G.; Zhang, M.-L.; and Sugiyama, M. 2024.
\newblock Binary classification with confidence difference.
\newblock \emph{Advances in Neural Information Processing Systems}, 36.

\bibitem[{Weber et~al.(2020)Weber, Hu{\ss}mann, Han, Matthes, and Liu}]{weber2020draw}
Weber, T.; Hu{\ss}mann, H.; Han, Z.; Matthes, S.; and Liu, Y. 2020.
\newblock Draw with me: Human-in-the-loop for image restoration.
\newblock In \emph{Proceedings of the 25th International Conference on Intelligent User Interfaces}, 243--253.

\bibitem[{Wei et~al.(2020)Wei, Feng, Chen, and An}]{wei2020combating}
Wei, H.; Feng, L.; Chen, X.; and An, B. 2020.
\newblock Combating noisy labels by agreement: A joint training method with co-regularization.
\newblock In \emph{Proceedings of the IEEE/CVF conference on computer vision and pattern recognition}, 13726--13735.

\bibitem[{Wei et~al.(2022{\natexlab{a}})Wei, Tao, Xie, Feng, and An}]{wei2022open}
Wei, H.; Tao, L.; Xie, R.; Feng, L.; and An, B. 2022{\natexlab{a}}.
\newblock Open-sampling: Exploring out-of-distribution data for re-balancing long-tailed datasets.
\newblock In \emph{International Conference on Machine Learning}, 23615--23630. PMLR.

\bibitem[{Wei et~al.(2022{\natexlab{b}})Wei, Liu, Liu, Niu, Sugiyama, and Liu}]{wei2022smooth}
Wei, J.; Liu, H.; Liu, T.; Niu, G.; Sugiyama, M.; and Liu, Y. 2022{\natexlab{b}}.
\newblock To Smooth or Not? When Label Smoothing Meets Noisy Labels.
\newblock In \emph{International Conference on Machine Learning}, 23589--23614. PMLR.

\bibitem[{Wei and Liu(2021)}]{wei2021when}
Wei, J.; and Liu, Y. 2021.
\newblock When Optimizing f-Divergence is Robust with Label Noise.
\newblock In \emph{International Conference on Learning Representations}.

\bibitem[{Wei et~al.(2022{\natexlab{c}})Wei, Zhu, Cheng, Liu, Niu, and Liu}]{wei2022learning}
Wei, J.; Zhu, Z.; Cheng, H.; Liu, T.; Niu, G.; and Liu, Y. 2022{\natexlab{c}}.
\newblock Learning with Noisy Labels Revisited: A Study Using Real-World Human Annotations.
\newblock In \emph{International Conference on Learning Representations}.

\bibitem[{Wei et~al.(2023)Wei, Zhu, Luo, Amid, Kumar, and Liu}]{wei2023aggregate}
Wei, J.; Zhu, Z.; Luo, T.; Amid, E.; Kumar, A.; and Liu, Y. 2023.
\newblock To aggregate or not? learning with separate noisy labels.
\newblock In \emph{Proceedings of the 29th ACM SIGKDD Conference on Knowledge Discovery and Data Mining}, 2523--2535.

\bibitem[{Wu et~al.(2021)Wu, Xiao, Sun, Zhang, Ma, and He}]{wu2021survey}
Wu, X.; Xiao, L.; Sun, Y.; Zhang, J.; Ma, T.; and He, L. 2021.
\newblock A Survey of Human-in-the-loop for Machine Learning.
\newblock \emph{arXiv preprint arXiv:2108.00941}.

\bibitem[{Xia et~al.(2023)Xia, Lu, Gong, Han, Yu, and Liu}]{xia2023regularly}
Xia, X.; Lu, P.; Gong, C.; Han, B.; Yu, J.; and Liu, T. 2023.
\newblock Regularly truncated m-estimators for learning with noisy labels.
\newblock \emph{IEEE Transactions on Pattern Analysis and Machine Intelligence}.

\bibitem[{Xiao et~al.(2015)Xiao, Xia, Yang, Huang, and Wang}]{xiao2015learning}
Xiao, T.; Xia, T.; Yang, Y.; Huang, C.; and Wang, X. 2015.
\newblock Learning from massive noisy labeled data for image classification.
\newblock In \emph{Proceedings of the IEEE conference on computer vision and pattern recognition}, 2691--2699.

\bibitem[{Yamins et~al.(2014)Yamins, Hong, Cadieu, Solomon, Seibert, and DiCarlo}]{yamins2014performance}
Yamins, D.~L.; Hong, H.; Cadieu, C.~F.; Solomon, E.~A.; Seibert, D.; and DiCarlo, J.~J. 2014.
\newblock Performance-optimized hierarchical models predict neural responses in higher visual cortex.
\newblock \emph{Proceedings of the national academy of sciences}, 111(23): 8619--8624.

\bibitem[{Yang and Alonso(2024)}]{Yang2024}
Yang, D.; and Alonso, O. 2024.
\newblock A bespoke question intent taxonomy for e-commerce.
\newblock In \emph{SIGIR 2024 eCommerce}.

\bibitem[{Yang et~al.(2024)Yang, Chen, Rao, Guo, Zhang, Yang, and Zhang}]{yang2024tackling}
Yang, D.; Chen, K.; Rao, J.; Guo, X.; Zhang, Y.; Yang, J.; and Zhang, Y. 2024.
\newblock Tackling vision language tasks through learning inner monologues.
\newblock In \emph{Proceedings of the AAAI Conference on Artificial Intelligence}, volume~38, 19350--19358.

\bibitem[{Yang et~al.(2017)Yang, Zhang, Chen, Zhang, and Chen}]{yang2017suggestive}
Yang, L.; Zhang, Y.; Chen, J.; Zhang, S.; and Chen, D.~Z. 2017.
\newblock Suggestive annotation: A deep active learning framework for biomedical image segmentation.
\newblock In \emph{International conference on medical image computing and computer-assisted intervention}, 399--407. Springer.

\bibitem[{Yao et~al.(2012)Yao, Gall, Leistner, and Van~Gool}]{yao2012interactive}
Yao, A.; Gall, J.; Leistner, C.; and Van~Gool, L. 2012.
\newblock Interactive object detection.
\newblock In \emph{2012 IEEE conference on computer vision and pattern recognition}, 3242--3249. IEEE.

\bibitem[{Yu et~al.(2019)Yu, Han, Yao, Niu, Tsang, and Sugiyama}]{yu2019does}
Yu, X.; Han, B.; Yao, J.; Niu, G.; Tsang, I.; and Sugiyama, M. 2019.
\newblock How does disagreement help generalization against label corruption?
\newblock In \emph{International Conference on Machine Learning}, 7164--7173. PMLR.

\bibitem[{Zhang and Sabuncu(2018)}]{zhang2018generalized}
Zhang, Z.; and Sabuncu, M. 2018.
\newblock Generalized cross entropy loss for training deep neural networks with noisy labels.
\newblock \emph{Advances in neural information processing systems}, 31.

\bibitem[{Zhang et~al.(2019)Zhang, Singh, Gadiraju, and Anand}]{zhang2019dissonance}
Zhang, Z.; Singh, J.; Gadiraju, U.; and Anand, A. 2019.
\newblock Dissonance between human and machine understanding.
\newblock \emph{Proceedings of the ACM on Human-Computer Interaction}, 3(CSCW): 1--23.

\bibitem[{Zhou et~al.(2018)Zhou, Bau, Oliva, and Torralba}]{zhou2018interpreting}
Zhou, B.; Bau, D.; Oliva, A.; and Torralba, A. 2018.
\newblock Interpreting deep visual representations via network dissection.
\newblock \emph{IEEE transactions on pattern analysis and machine intelligence}, 41(9): 2131--2145.

\bibitem[{Zhu, Liu, and Liu(2021)}]{zhu2021second}
Zhu, Z.; Liu, T.; and Liu, Y. 2021.
\newblock A second-order approach to learning with instance-dependent label noise.
\newblock In \emph{Proceedings of the IEEE/CVF conference on computer vision and pattern recognition}, 10113--10123.

\bibitem[{Zimmermann et~al.(2021)Zimmermann, Borowski, Geirhos, Bethge, Wallis, and Brendel}]{zimmermann2021well}
Zimmermann, R.~S.; Borowski, J.; Geirhos, R.; Bethge, M.; Wallis, T.; and Brendel, W. 2021.
\newblock How Well do Feature Visualizations Support Causal Understanding of {CNN} Activations?
\newblock \emph{Advances in Neural Information Processing Systems}, 34: 11730--11744.

\end{thebibliography}
